\documentclass{vislang_lab_2026}

\usepackage{graphicx}
\usepackage{booktabs}
\usepackage{amsfonts}
\usepackage{amsmath}
\usepackage{nicefrac}
\usepackage{xcolor}
\usepackage{subcaption}
\usepackage{algorithm}
\usepackage{algorithmic}
\usepackage{multirow}
\usepackage{wrapfig}
\usepackage{makecell}
\usepackage{orcidlink}
\usepackage{tabularx}
\usepackage{cleveref}
\usepackage{tcolorbox}
\tcbuselibrary{breakable,listings}
\usepackage{silence}
\newtcolorbox{agentinput}[1][]{%
    breakable,
    colback=blue!5!white,
    colframe=blue!70!black,
    fonttitle=\bfseries,
    title=Input,
    boxrule=0.8pt,
    arc=1mm,
    left=1mm,right=1mm,top=1mm,bottom=1mm,
    #1
}

\newtcolorbox{agentoutput}[1][]{%
    breakable,
    colback=green!5!white,
    colframe=green!60!black,
    fonttitle=\bfseries,
    title=Output,
    boxrule=0.8pt,
    arc=1mm,
    left=1mm,right=1mm,top=1mm,bottom=1mm,
    #1
}

\newtcolorbox{agentbehavior}[1][]{
    breakable,
    colback=gray!5!white,
    colframe=gray!60!black,
    fonttitle=\bfseries,
    title=Behavior,
    boxrule=0.8pt,
    arc=1mm,
    left=1mm,right=1mm,top=1mm,bottom=1mm,
    #1
}

\newtcolorbox{systemprompt}[1][]{
    colback=blue!5!white,
    colframe=blue!70!black,
    fonttitle=\bfseries,
    breakable,
    #1
}

\newtcolorbox{userprompt}[1][]{%
    colback=green!5!white,
    colframe=green!60!black,
    fonttitle=\bfseries,
    breakable,
    #1
}

\newtcolorbox{ideabox}[1][]{
    colback=orange!5!white,
    colframe=orange!70!black,
    fonttitle=\bfseries,
    breakable,
    #1
}

\newtcblisting{codebox}[1][]{
    colback=gray!3!white,
    colframe=gray!50!black,
    fonttitle=\bfseries,
    breakable,
    listing only,
    listing options={
        language=Python,
        basicstyle=\scriptsize\ttfamily,
        breaklines=true,
        tabsize=4,
        showstringspaces=false,
        columns=fullflexible,
        keepspaces=true,
    },
    #1
}

\title{Agentic Discovery with Active Hypothesis Exploration for Visual Recognition}

\author[1]{Jaywon Koo}
\author[1]{Jefferson Hernandez}
\author[1]{Ruozhen He}
\author[1]{Hanjie Chen}
\author[1]{Chen Wei}
\author[1]{Vicente Ordonez}

\affil[1]{Rice University}

\newcommand{\method}[0]{HypoExplore}

\begin{document}

\maketitle

\begin{abstract}
We introduce \method{}, an agentic framework that formulates neural architecture discovery for visual recognition as a hypothesis-driven scientific inquiry. Given a human-specified high-level research direction, \method{} ideates, implements, evaluates, and improves neural architectures through evolutionary branching.  New hypotheses are created using a large language model by selecting a parent hypothesis to build upon, guided by a dual strategy that balances exploiting validated principles with resolving uncertain ones.
Our proposed framework maintains a Trajectory Tree that records the lineage of all proposed architectures, and a Hypothesis Memory Bank that actively tracks confidence scores acquired through experimental evidence. 
After each experiment, multiple feedback agents analyze the results from different perspectives and consolidate their findings into hypothesis confidence updates. Our framework is tested on discovering lightweight vision architectures on CIFAR-10, with the best achieving 94.11\% accuracy evolved from a root node baseline that starts at 18.91\%, and generalizes to CIFAR-100 and Tiny-ImageNet. We further demonstrate applicability to a specialized domain by conducting independent architecture discovery runs on MedMNIST, which yield a state-of-the-art performance.
We show that hypothesis confidence scores grow increasingly predictive as evidence accumulates, and that the learned principles transfer across independent evolutionary lineages, suggesting that \method{} not only discovers stronger architectures, but can help build a genuine understanding of the design space. 
\end{abstract}
\section{Introduction}
\label{sec:intro}
Designing effective neural architectures remains a central challenge in computer vision. Despite the success of modern deep learning and our advanced understanding of how to design and engineer architectures for standard benchmarks~\cite{kirillov2023segment, ravi2024sam, liu2021swin, caron2021emerging}, discovering strong architectures for specialized domains still requires substantial human effort, repeated experimentation, and careful iteration. At the same time, recent advances in large language models and multi-agent systems~\cite{si2026towards, chen2026mars, cheng2025language} have made it increasingly feasible to automate parts of this process, including code generation~\cite{copet2025cwm, weng2026groupevolvingagentsopenendedselfimprovement, zhang2025darwingodelmachineopenended}, experiment execution~\cite{huang2023mlagentbench, si2026towards}, debugging~\cite{epperson2025interactive}, and result analysis~\cite{koo2024proptest}. These developments suggest the possibility of autonomous systems that can assist with, and potentially accelerate, neural architecture discovery~\cite{wen2020neural, ren2021comprehensive, cheng2025language} and beyond.

\begin{figure}[t]
  \centering
  \includegraphics[width=0.95\linewidth]{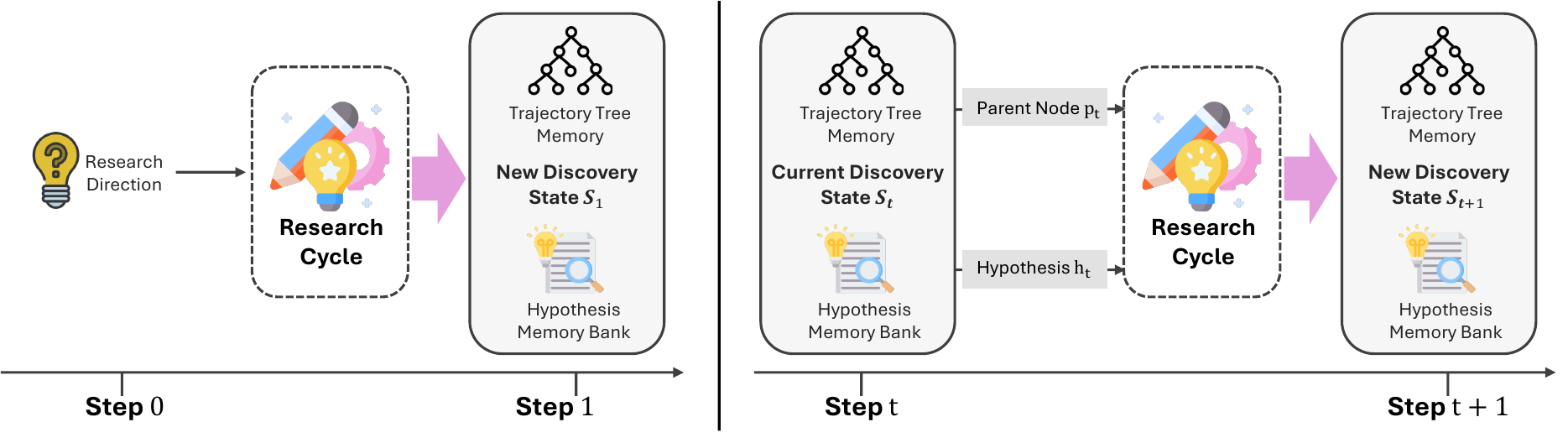}
  \caption{\textbf{High-level Overview of \method{}.} Starting from a research direction, \method{} initializes a discovery state with a Trajectory Tree Memory and Hypothesis Memory Bank (Step 0$\rightarrow$ Step 1). At each subsequent step, the current discovery state selects a parent node and hypothesis to guide the Research Cycle, producing an updated discovery state with enriched memory (Step t $\rightarrow$ Step t+1). }
  \label{fig:overview}
  \vspace{-0.1in}
\end{figure}

Recently proposed frameworks in automated architecture discovery and experimentation have demonstrated that they can successfully generate and iterate over implementations and execute experiments efficiently~\cite{yang2025nader, chang2025revonad, si2026towards, liu2025alphago,yu2025alpharesearchacceleratingnewalgorithm}. These methods often explore the design space through targeted architectural modifications, improved design patterns, and hyperparameter tuning. Our work aims to go further by conducting broader from-scratch discovery that can avoid falling into repeated design patterns and overly constrained local modifications by relying on more explicit hypothesis tracking and formulation. Our goal is to build a system that is effective at \emph{running} experiments, but more principled in deciding \emph{which} research direction to pursue next. Our proposal aims to diminish the chances of exploration becoming myopic, redundant, and difficult to interpret.

In this work, we argue that automated neural architecture discovery should be framed not merely as architecture search, but as a process of \emph{autonomous scientific discovery}. 
Recent LLM-based neural architecture design systems~\cite{yang2025nader, chang2025revonad} have already moved beyond fixed search spaces which has allowed significantly more exploration that goes well beyond hyperparameter tuning or improvements on optimization. 
Our proposed \method{} framework further promotes exploration by not using a predefined seed architecture as the starting point. By  not anchoring our exploration to a fixed initial design, we aim to depart from incremental updates and refinements. We posit that the deeper challenge of autonomous discovery is deciding what fundamentally new architectural idea to pursue next, and on what evidential basis.

Meta-research on scientific practice provides exactly this foundation. It characterizes discovery as a coupled search over a \emph{hypothesis space} and an \emph{experiment space}, where progress depends on managing the interaction between proposing explanations and selecting informative tests~\cite{klahr1988dual}.
It further emphasizes maintaining \emph{multiple competing hypotheses} to avoid fixation and redundancy, and prioritizing tests that can eliminate alternatives rather than merely accumulate confirmations~\cite{chamberlin1890method, platt1964strong, wason1960failure}.
Similarly, work in organizational learning and the sociology of science highlights the exploration--exploitation tension and the benefits of division of cognitive labor, motivating structured mechanisms for allocating effort across promising directions while still probing uncertain ones~\cite{march1991exploration, kitcher1990division}. 
Together, these insights motivate an architecture discovery system grounded not in arbitrary generation, but in explicit, evidence-driven hypothesis management.

\begin{wrapfigure}{r}{2.2in}
\vspace{-0.18in}
    \centering
    \includegraphics[width=\linewidth]{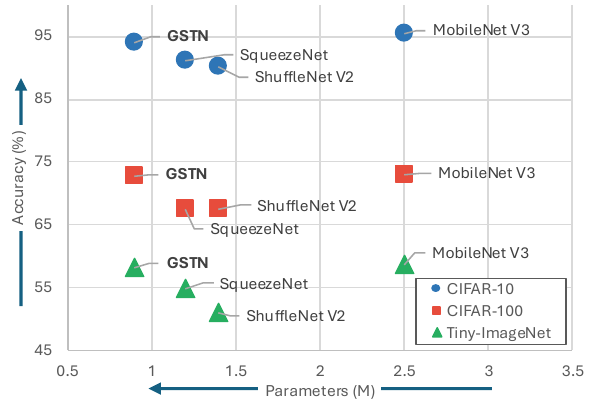}
    \caption{\method{} finds a lightweight Global Shape Token Network (GSTN) that introduces a small bank of learned global vectors. This network using less parameters closely matches or surpasses other manually engineered networks. }
    \label{fig:results-summary}
    \vspace{-0.25in}
\end{wrapfigure}
Accordingly, instead of treating candidate models as isolated architecture instances, we represent each design direction as an explicit architectural hypothesis: a structured conjecture about what kind of mechanism may improve performance.
This perspective shifts the role of the system from simply proposing architectural variants to managing an iterative scientific process, including generating hypotheses, filtering redundant proposals, implementing selected ideas, evaluating them empirically, and refining future decisions using accumulated evidence.
By making hypotheses explicit, the discovery process becomes more structured, less repetitive, and more interpretable.

\method{} is a memory-grounded multi-agent framework for autonomous scientific discovery of neural architectures. \method{} starts from a human-specified research direction rather than a predefined seed architecture, and improves itself through iterative hypothesis testing and feedback-driven memory updates (Figure~\ref{fig:overview}). The framework contains specialized agents for idea generation, redundancy filtering, code implementation, experiment execution, and feedback analysis. Its memory system has two complementary components. First, a trajectory tree stores complete research branches, including hypotheses, implementations, and observed outcomes, preserving the full history of exploration. Second, a hypothesis memory bank tracks hypothesis usage, testing logs, and confidence estimates, enabling the system to avoid repeated trials and reason about which directions remain promising.

Building on this memory, we further propose a dual selection strategy that guides exploration at two levels. A parent-node selector determines which research branch to expand by considering both empirical promise and remaining unexplored potential. A hypothesis selector then chooses which candidate hypothesis to evaluate next by balancing exploitation of high-confidence directions with exploration under uncertainty. Together, these mechanisms allow \method{} to conduct more deliberate and interpretable discovery than a simple loop of generation and execution.
Our contributions are summarized as follows:
\vspace{-0.1in}
\begin{itemize}
    \item We introduce \method{}, a memory-grounded multi-agent framework that formulates automated neural architecture design from scratch for autonomous scientific discovery.
    \item We propose an explicit hypothesis-centered memory system, consisting of a trajectory tree and a hypothesis memory bank, to support non-redundant and interpretable exploration.
    \item We develop a selection strategy over research branches and candidate hypotheses, enabling the balance of empirical promise with unexplored potential.
    \item We demonstrate that \method{} discovers an efficient architecture reaching 94.11\% on CIFAR-10, generalizes robustly to CIFAR-100 and Tiny-ImageNet (Figure \ref{fig:results-summary}), and through an independent discovery run on MedMNIST achieves state-of-the-art performance, establishing applicability across both general and domain-specific visual recognition.
\end{itemize}
\section{Related Work}
\label{sec:related}

\vspace{-0.05in}
\subsection{Autonomous Scientific Discovery (ASD).}
Recent ASD systems use LLMs to close the loop between ideation, implementation, execution, and reflection, but differ in what drives exploration and what the ``discovered object'' is. 
AutoDiscovery studies open-ended discovery using Bayesian surprise as an intrinsic reward and MCTS-style search over nested hypotheses~\cite{agarwal2025autodiscovery}. MARS instead targets automated AI research using practices from \emph{SWE} and reflective memory across branches~\cite{chen2026mars}. 
\emph{Genesys} simulates the research lifecycle for discovering language modeling architectures, using genetic programming and and tight execution budgets
~\cite{cheng2025language}. Other execution-grounded research agents and AI-scientist frameworks similarly emphasize code execution, reflection, and memory for recipe-level, repository-level, or cross-domain analyses~\cite{si2026towards,yang2025rdagentllmagentframeworkautonomous,yu2025alpharesearchacceleratingnewalgorithm,mitchener2025kosmosaiscientistautonomous,yu2025tinyscientistinteractiveextensiblecontrollable}. In contrast, our setting is autonomous \emph{neural architecture} discovery for vision, where the search object is an evolving architecture lineage. \method{} therefore makes architectural hypotheses explicit and uses branch-level and hypothesis-level memory to decide which lineage to expand and which uncertain mechanism to test next.

\subsection{Hypothesis Generation and Evaluation.}
A complementary line studies literature-grounded hypothesis generation and theory construction. ResearchAgent generates and refines research ideas from scientific literature, while BioDisco, MOOSE-Chem, and HypER emphasize evidence-grounded hypothesis generation via knowledge graphs, inspiration retrieval, or provenance-aware reasoning chains~\cite{baek2025researchagentiterativeresearchidea,ke2025biodiscomultiagenthypothesisgeneration,yang2025moosechemlargelanguagemodels,vasu-etal-2025-hyper}. Recent systems also synthesize higher-level scientific theories or validate free-form hypotheses through agentic falsification, statistical evidence aggregation, or uncertainty-aware refinement~\cite{jansen2026generatingliteraturedrivenscientifictheories,huang2025automatedhypothesisvalidationagentic,duan2025bayesentropycollaborativedrivenagents}. Unlike these methods, our hypotheses are not final output in text form: each hypothesis in \method{} is an actionable architectural mechanism instantiated as runnable model code, evaluated on the target task, and written back into structured discovery memory.

\subsection{Self-Evolving Agents and Memory-Augmented Improvement. }
Another related direction focuses on improving the \emph{researcher} itself. Self-evolving coding agents such as Darwin G\"odel Machine, Group-Evolving Agents, and AlphaEvolve iteratively modify agent code or executable programs and retain strong variants through open-ended evolution~\cite{zhang2025darwingodelmachineopenended,weng2026groupevolvingagentsopenendedselfimprovement,novikov2025alphaevolvecodingagentscientific}. In parallel, memory-centric methods such as ReasoningBank, Dynamic Cheatsheet, and Agentic Context Engineering distill reusable reasoning strategies, snippets, or evolving contexts from prior trajectories to improve future performance~\cite{ouyang2025reasoningbankscalingagentselfevolving,suzgun2025dynamiccheatsheettesttimelearning,zhang2026agenticcontextengineeringevolving}. Our goal is different: we keep the discovery framework fixed and evolve the \emph{discovered artifact}, namely the architecture lineage. Accordingly, our memory stores branch histories and per-hypothesis evidence rather than generic reasoning traces or prompt playbooks.

\subsection{Neural Architecture Design (NAD).}
The closest line of work is LLM-based neural architecture design. NADER formulates architecture design as multi-agent collaboration and uses reflection together with graph-based architecture representations to reduce repeated mistakes and code-generation noise~\cite{yang2025nader}. RevoNAD combines multi-expert consensus, reflective exploration, and Pareto-guided evolutionary selection to encourage diverse and deployable architectures~\cite{chang2025revonad}. Our method shares the goal of moving beyond fixed search spaces, but differs in how exploration is organized. Rather than relying primarily on reflective editing or population-level evolutionary orchestration, \method{} performs from-scratch discovery around explicit architectural hypotheses, a trajectory tree that records lineage, and a hypothesis memory bank that accumulates reusable evidence. This yields a dual decision process over \emph{where} to expand and \emph{what} to test, making discovery more structured, interpretable, and less redundant.
\section{Method}
\label{sec:method}

\method{} is a hypothesis-grounded multi-agent framework for autonomous scientific discovery of neural architectures. As shown in Figure~\ref{fig:overview},
\method{} operates in a predefined task domain (image classification in this paper). Given a human-specified research agenda, the system aims to discover effective neural architectures \emph{from scratch}, without a seed backbone or a fixed search space.

\paragraph{Overview.}
\method{} maintains a discovery state $\mathcal{S}_t = \{\mathcal{T}_t, \mathcal{M}_t\}$, where $\mathcal{T}_t$ records the experimental lineage and $\mathcal{M}_t$ stores hypothesis-level statistics. At iteration $t$, it selects a parent node $p_t$, selects a small \emph{set} of hypotheses $\mathcal{Q}_t$ to test under that parent, and runs a research cycle to instantiate, execute, and analyze architectures conditioned on $(p_t, \mathcal{Q}_t)$. The resulting outcomes update both $\mathcal{T}_t$ and $\mathcal{M}_t$, yielding $\mathcal{S}_{t+1}$.
Rather than mutating architectures within a predefined family, \method{} performs iterative hypothesis-driven discovery, using structured memory to decide \emph{where} to explore next and \emph{what} mechanisms to test. We describe the memory (Sec.~\ref{sec:memory}), research cycle (Sec.~\ref{sec:research_cycle}), and dual selection mechanism (Sec.~\ref{sec:dual_selection}).

\subsection{Structured Memory}
\label{sec:memory}

The discovery state $\mathcal{S}_t = \{\mathcal{T}_t, \mathcal{M}_t\}$ is a structured memory with two complementary stores. This separation lets the system reason both about \emph{branch trajectories} (what lines of exploration are promising) and \emph{hypotheses} (what mechanistic claims are supported or contradicted across experiments).

\para{Trajectoty tree.}
The trajectory tree $\mathcal{T}_t$ records the branching structure of discovery. Each node $v \in \mathcal{T}_t$ corresponds to one executed research step and stores
$v = \{h_v, a_v, r_v, p_v\}$, where $h_v$ is the architectural hypothesis (or hypothesis set reference) used to guide the design, $a_v$ is the instantiated architecture, $r_v$ is the experimental outcome, and $p_v$ is the parent node.
By preserving parent-child relations, $\mathcal{T}_t$ exposes full exploration trajectories, enabling the system to identify promising, saturated, or repeatedly failing branches.

\para{Hypothesis memory bank.}
The hypothesis memory bank $\mathcal{M}_t$ aggregates statistics across related hypotheses. For each hypothesis $h$, the bank maintains
$m_t(h)=\{N_t(h),\, c_t(h),\, \ell_t(h)\}$, where $N_t(h)$ is the number of times $h$ has been tested, $c_t(h)\in[0,1]$ is its current confidence score, and $\ell_t(h)$ stores logs such as supporting evidence, contradictions, failure modes, and implementation notes.

\begin{figure}[t]
  \centering
  \includegraphics[width=\linewidth]{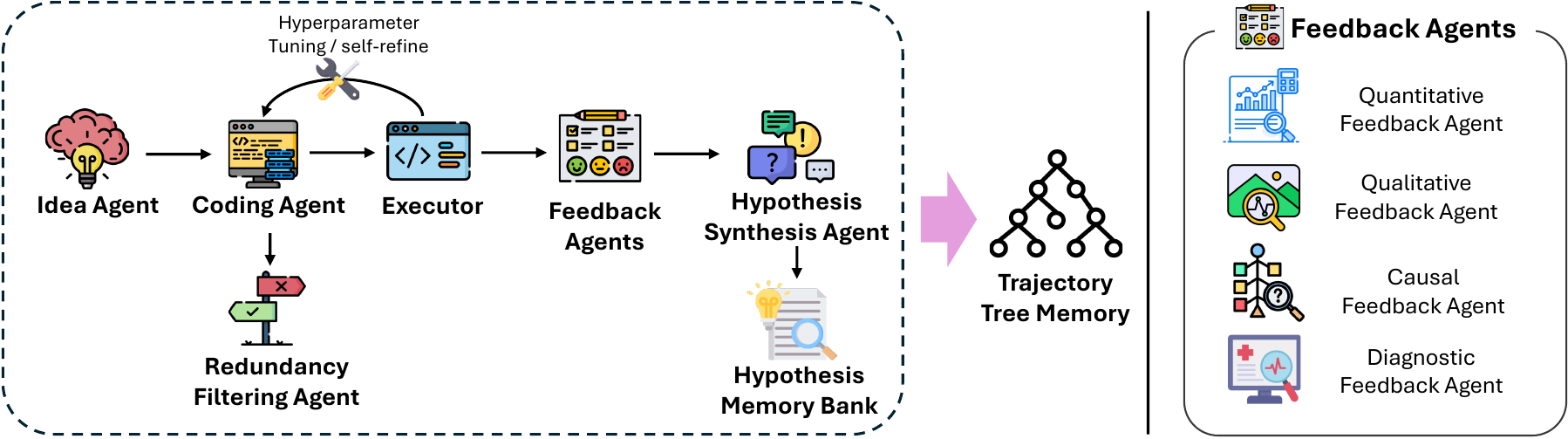}
  \caption{\textbf{Overview of Per-Node Research Cycle.} The Idea Agent proposes a neural architecture, which the Coding Agent implements with iterative hyperparameter tuning. A Redundancy Filtering Agent checks against the Tree Memory to prevent re-generation of concepts already explored. The Executor trains and evaluates each architecture, and the results are analyzed by four specialized Feedback Agents (right), each providing a distinct analytical perspective. The Hypothesis Synthesis Agent consolidates these multi-perspective analyses to update the Hypothesis Memory Bank, while experimental trajectories are stored in the Tree Memory to guide future exploration.}
  \label{fig:pipeline}
\end{figure}

\subsection{Per-Node Research Cycle}
\label{sec:research_cycle}

Each iteration executes a pipeline of specialized agents (Figure~\ref{fig:pipeline}).
Given a parent node $p_t$ and a selected hypothesis set $\mathcal{Q}_t$, the system attempts to create up to $|\mathcal{Q}_t|$ child nodes by running the cycle once per selected hypothesis. If multiple hypotheses lead to duplicate proposals, redundancy filtering rejects and regenerates until novelty is satisfied or the retry budget is exhausted.

\para{Idea Agent.}
The Idea Agent receives the research direction, parent context (architecture specification, performance, and multi-agent feedback), and the hypothesis memory $\mathcal{M}_t$.
In \emph{root node} (generation~0), it generates architectures from scratch guided by the research direction.
In \emph{evolution mode}, it conditions on a selected hypothesis $h \in \mathcal{Q}_t$ and the parent node $p_t$ to produce: (i) an architecture specification, (ii) a reasoning trace, (iii) references to existing hypotheses in $\mathcal{M}_t$, and (iv) up to $K_{\mathrm{hyp}}$ newly proposed hypotheses motivated by the design.

\para{Coding Agent.}
The Coding Agent receives the architecture specification together with implementation notes from $\ell_t(h)$ (failure modes and recommended practices from prior experiments).
It generates \texttt{model.py} and \texttt{config.py}. An error recovery loop reruns the agent with the error trace for up to $R_{\max}$ attempts.
After successful compilation, a hyperparameter refinement loop adjusts only \texttt{config.py} for up to $F_{\max}$ steps, with early stopping on accuracy plateau.

\para{Redundancy filtering.}
Before execution, an LLM judge compares the proposed architecture against the top-$k$ most similar archived concepts in $\mathcal{T}_t$. If the architecture is judged to be a duplicate, the node is rejected and the pipeline backtracks to regenerate. This enforces novelty as a hard constraint on node creation.

\para{Executor.}
The Executor trains each proposed architecture $a$ under a wall-clock timeout $\tau_{\max}$. A sanity-check phase (5 epochs) detects catastrophic failures early.
The recorded outcome is $r = \{J(a), d\}$, where $J(a)$ is task performance and $d$ stores diagnostics (e.g., instability, timeout). Each successful execution appends a new node to the trajectory tree, storing $(h, a, r)$ and its parent pointer.

\newpage
\para{Multi-perspective feedback.}
On success, four parallel agents analyze the outcome from complementary perspectives:
\emph{(i)~Quantitative}: analyzes accuracy, loss curves, convergence speed, and computational efficiency, extracting hypothesis evidence from performance patterns.
\emph{(ii)~Qualitative}: a VLM examines misclassified images and attention maps.
\emph{(iii)~Causal}: compares the parent and child architectures, attributing observed performance changes to specific structural modifications.
\emph{(iv)~Diagnostic} (failure/timeout only): performs root-cause analysis and records implementation failure modes into $\ell_t(h)$.

\para{Hypothesis synthesis.} A hypothesis synthesis agent consolidates feedbacks in one LLM call, deduplicating overlapping updates, resolving disagreements in evidence interpretation, and capping new hypotheses at $K_{\mathrm{synth}}$ per node. Each proposed hypothesis must pass a quality gate assessing mechanistic specificity, falsifiability, novelty w.r.t.\ $\mathcal{M}_t$, and actionability before admission to the memory bank.

\para{Memory update.}
Confidence scores of all referenced hypotheses are updated using evidence type and strength $w \in [0,1]$ produced by hypothesis synthesis:
\begin{equation}
c_{t+1}(h) =
\begin{cases}
c_t(h) + \eta\, w\,(1 - c_t(h)) & \text{if evidence supports } h,\\
c_t(h) - \eta\, w\, c_t(h) & \text{if evidence contradicts } h,
\end{cases}
\label{eq:confidence_update}
\end{equation}
where $\eta \in (0,1]$ is a learning rate. The factors $(1-c_t(h))$ and $c_t(h)$ keep confidence bounded: supporting evidence pushes confidence toward~$1$ with diminishing increments, while contradicting evidence pushes it toward~$0$.
Hypotheses are initialized at $c_0(h)=0.5$, representing maximum uncertainty. Finally, the hypothesis logs $\ell_t(h)$ and counts $N_t(h)$ are updated with newly synthesized evidence, failure modes, and references.

\subsection{Dual Selection for Guided Discovery}
\label{sec:dual_selection}
To avoid undirected trial-and-error, we use a two-stage selection strategy that separates \emph{which branch} to expand from \emph{which hypotheses} to test within that branch. We keep parent selection deterministic for stability at the branch level, and concentrate the exploration--exploitation trade-off in hypothesis selection.

\subsubsection{Parent-Node Selection. }
\label{sec:parent_selection}

Let $\mathcal{V}_t^{\mathrm{exp}}$ denote the set of expandable nodes in the trajectory tree at iteration $t$. For each candidate node $n \in \mathcal{V}_t^{\mathrm{exp}}$, we compute a branch quality score by combining task performance and execution efficiency:
\begin{equation}
\mathrm{quality}(n)
=
\lambda_{\mathrm{acc}}\, \mathrm{Acc}(n)
+
(1-\lambda_{\mathrm{acc}})
\left(
1 - \frac{\min(\tau(n), \tau_{\max})}{\tau_{\max}}
\right),
\label{eq:quality}
\end{equation}
where $\mathrm{Acc}(n) \in [0,1]$ is normalized validation accuracy at node $n$, $\tau(n)$ is the observed training time, $\tau_{\max}$ is the maximum allowed training time, and $\lambda_{\mathrm{acc}} \in [0,1]$ controls the trade-off.
On top of quality, we measure whether the node still contains useful unexplored directions. Let $\mathcal{H}_{\mathrm{active}}(n)$ denote the set of active hypotheses associated with node $n$ (excluding confirmed and refuted hypotheses), and let $\mathcal{H}_{\mathrm{tested}}(n) \subseteq \mathcal{H}_{\mathrm{active}}(n)$ denote the subset already tested. We define an availability score
\begin{equation}
\mathrm{avail}(n)
= 1 - \frac{\left|\mathcal{H}_{\mathrm{tested}}(n)\right|}
{\left|\mathcal{H}_{\mathrm{active}}(n)\right|},
\label{eq:avail}
\end{equation}
with the convention $\mathrm{avail}(n)=0$ if $\left|\mathcal{H}_{\mathrm{active}}(n)\right|=0$.
The final parent score is a weighted combination:
\begin{equation}
\mathrm{score}(n)
=
\lambda_{\text{parent}}\,\mathrm{quality}(n)
+
\bigl(1-\lambda_{\text{parent}}\bigr)\,\mathrm{avail}(n),
\label{eq:parent_score}
\end{equation}
where $\lambda_{\mathrm{parent}} \in [0,1]$ trades-off branch quality and remaining search potential. We then select as parent the expandable 
node with the highest score.

\subsubsection{Hypothesis Selection.}
\label{sec:hypothesis_selection}

For node $p_t$, let $\mathcal{H}_{\mathrm{cand}}(p_t)$ denote the hypotheses not yet tested on its ancestors. We select an exploitation subset and an exploration subset from $\mathcal{H}_{\mathrm{cand}}(p_t)$; their union forms set $\mathcal{Q}_t$ passed to the research cycle.

\begin{figure*}[t]
    \centering
    \includegraphics[width=0.60\textwidth]{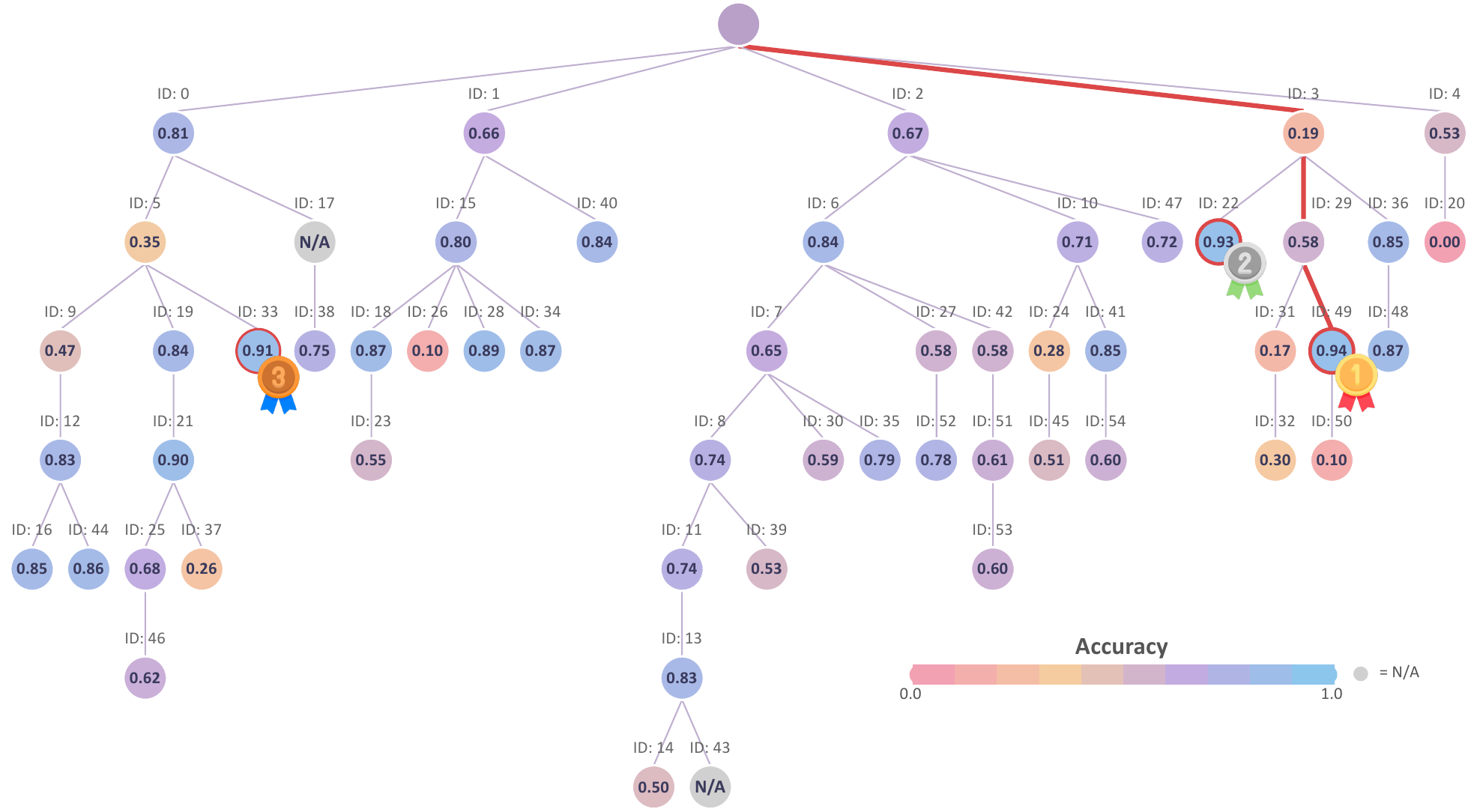}
    \hfill
    \includegraphics[width=0.38\textwidth]{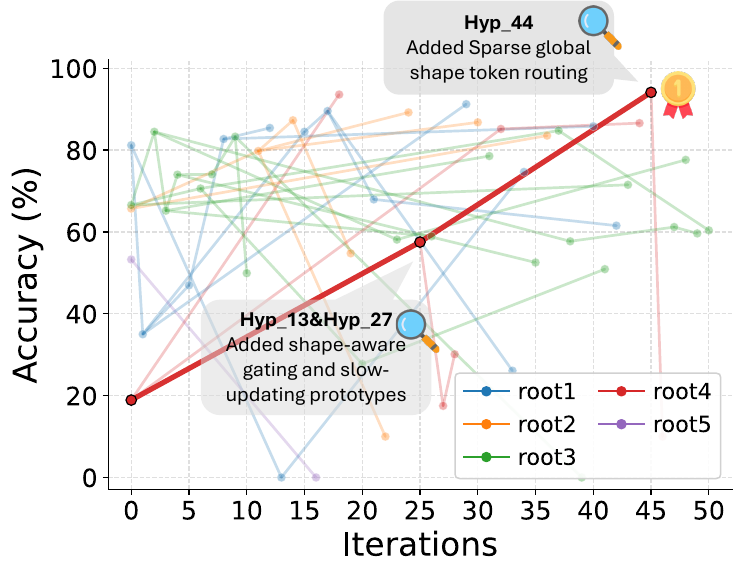}
    \caption{\textbf{\method{} discovers high-performing architectures via hypothesis-guided evolutionary branching.} The trajectory tree (left) records the full lineage of architectural experiments across 5 root branches, where node color indicates accuracy. While most branches improve gradually, one branch achieves a decisive accuracy leap by iteratively applying accumulated hypotheses, evolving from a poorly-performing root architecture to ultimately discover the best-performing design (right).
    }
    \label{fig:evolution}
\end{figure*}

\para{Exploitation via Thompson sampling.}
To avoid overcommitting to noisy early winners, we use Thompson sampling over weighted supporting and contradicting evidence~\cite{chapelle2011thompson,russo2018thompson}. For each $h \in \mathcal{H}_{\mathrm{cand}}(p_t)$, we define
\begin{equation}
\alpha_h
=
\alpha_0 + \sum_{e \in \mathcal{E}_{\mathrm{sup}}(h)} w_e,
\qquad
\beta_h
=
\beta_0 + \sum_{e \in \mathcal{E}_{\mathrm{con}}(h)} w_e,
\label{eq:beta_params}
\end{equation}
with prior pseudo-counts $\alpha_0,\beta_0 > 0$, and sample $\theta_h \sim \mathrm{Beta}(\alpha_h,\beta_h)$. Let $\pi$ order $\mathcal{H}_{\mathrm{cand}}(p_t)$ such that $\theta_{\pi(1)} \ge \theta_{\pi(2)} \ge \cdots$. The exploitation subset is $\mathcal{H}_{\mathrm{exploit}}^\star(p_t)=\{\pi(1), \ldots, \pi(K_{\mathrm{hypo}})\}.$

\para{Exploration via epistemic value.}
In parallel, we prioritize hypotheses whose current evidence is ambiguous. Using the confidence $c_t(h)\in[0,1]$ stored in $\mathcal{M}_t$, we define
\begin{equation}
\mathrm{epistemic}(h)
=
1 - \left|2\,c_t(h) - 1\right|,
\label{eq:epistemic}
\end{equation}
which is maximal at $c_t(h)=0.5$ and decreases toward $0$ or $1$. We then define $\mathcal{H}_{\mathrm{explore}}^\star(p_t)$ as the $K_{\mathrm{hypo}}$ hypotheses in $\mathcal{H}_{\mathrm{cand}}(p_t)$ with the largest $\mathrm{epistemic}(h)$, analogous to uncertainty-based acquisition in active learning~\cite{settles2009active,houlsby2011bald}.

\para{Final hypothesis set..}
We pass the deduplicated union
\[
\mathcal{Q}_t
=
\mathcal{H}_{\mathrm{exploit}}^\star(p_t)\cup \mathcal{H}_{\mathrm{explore}}^\star(p_t),
\]
so $|\mathcal{Q}_t| \le 2K_{\mathrm{hypo}}$. Together with parent selection, this yields branch-level continuation plus hypothesis-level exploitation and uncertainty-driven exploration.

\section{Experiment}
\label{sec:experiment}

\subsection{Experimental Setup}
We set our research direction as ``Design a novel attention mechanism where tokens influence each other through fundamentally different connection patterns than standard all-to-all self-attention'' and start by generating 5 root nodes.
Our agents are built on GPT-5-mini~\cite{openai_gpt5_for_developers_2025}. 
We set the parent-node quality weight to $\lambda_{\mathrm{acc}}=0.85,$ maximum allowed training time to $t_{\max}=30$ and $\lambda_{\mathrm{parent}}=0.60$ for parent-node ranking. 
For hypothesis exploitation, we use a uniform Beta prior with $(\alpha_0,\beta_0)=(1,1)$ and assign uniform evidence weights $w_e=1$. For memory update, we set the confidence update rate to $\eta_c=0.20$.  We set $K_{\mathrm{hypo}}=2$, making the dual selection stage to return $2K_{\mathrm{parent}}=4$ hypotheses per iteration. Baseline details, implementations, and evaluation protocols are in the Supplementary.

\subsection{Main Results}
\begin{figure*}[t]
    \centering
    \includegraphics[width=0.48\textwidth]{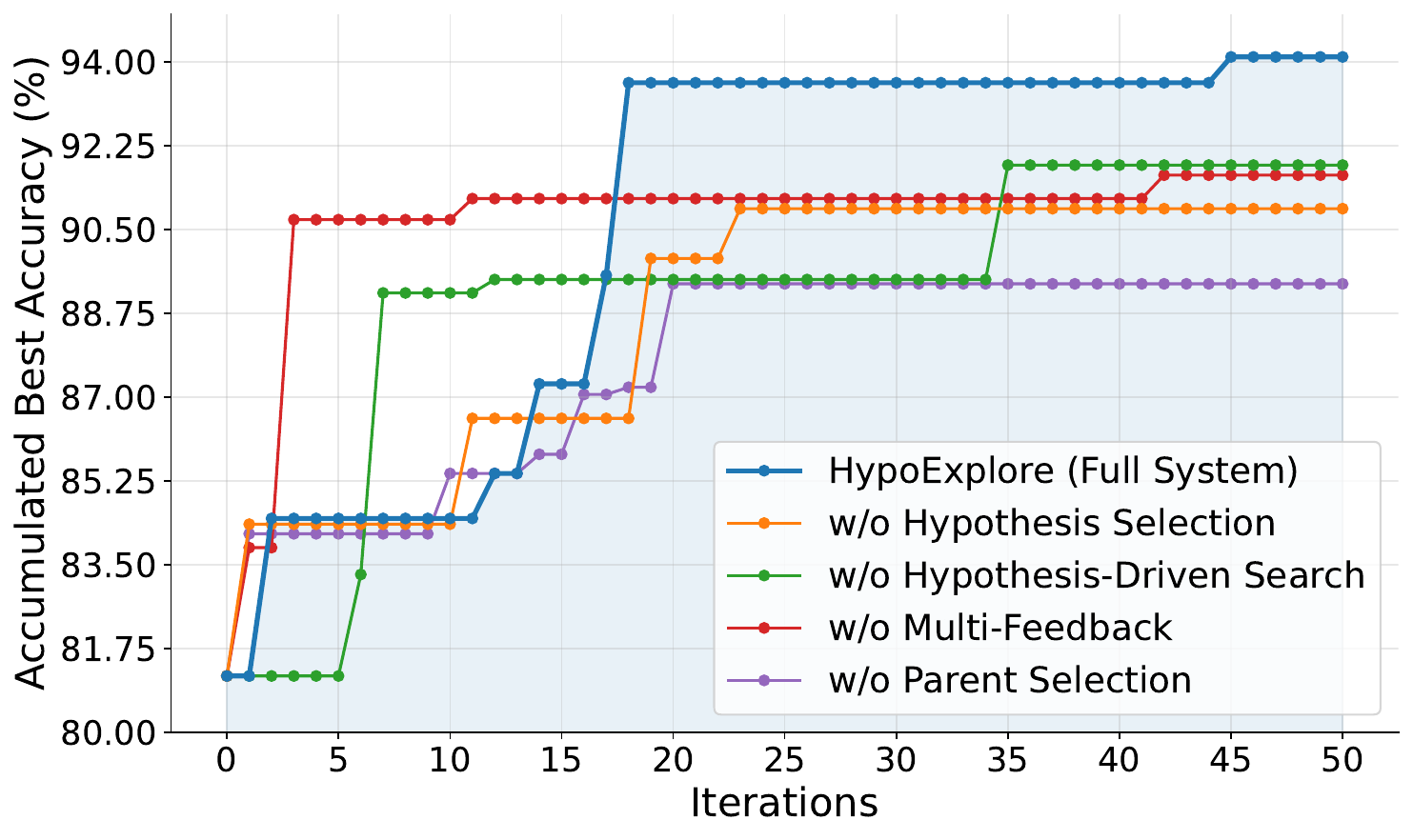}
    \hfill
    \includegraphics[width=0.48\textwidth]{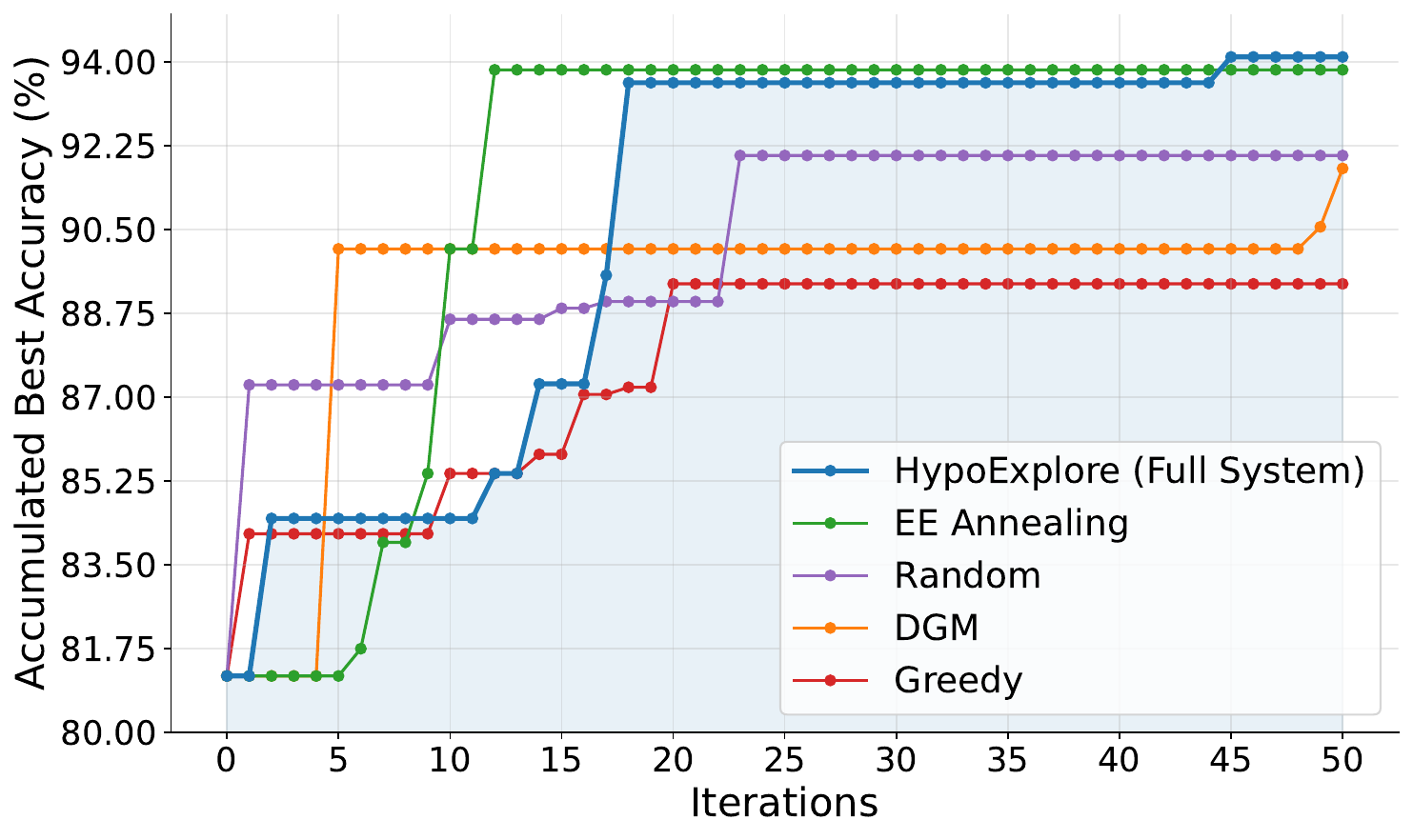}
    \caption{\textbf{Accumulated best accuracy over 50 iterations on CIFAR-10.} All methods share the same five root architectures (iteration 0, best 81.2\%). \textbf{Left:} Component ablation. Each curve removes one contribution from the full system. 
    \textbf{Right:} Parent selection strategy comparison. All methods have hypothesis memory, hypothesis selection, and multi-agent feedback, and only the parent selection mechanism varies. 
    }
    \label{fig:main_result}
\end{figure*}
Fig.~\ref{fig:evolution} visualizes the trajectory tree (left) and per-branch accuracy over 50 iterations (right), showing that branches follow distinct trajectories. Some improve steadily, while others decline before recovering. One branch eventually separates from the rest, driven by key hypotheses such as Hyp\_44, and yields the best-performing architecture.
Fig.~\ref{fig:main_result} tracks the best accuracy found up to each iteration. All methods start from the same five root architectures, with a maximum initial accuracy of 81.2\%. After 50 iterations, \method{} reaches 94.11\% accuracy on CIFAR-10 and improves steadily throughout the search: it reaches 81.28\% by iteration 15, surpasses 93.57\% by iteration 18, and continues improving to 94.11\% by iteration 45. This suggests that \method{} becomes increasingly effective over time rather than succeeding through a single fortunate discovery.

The full system outperforms variants that lack any one of its core components (Fig.~\ref{fig:main_result}, left). Without hypothesis-driven search (removing hypothesis memory and selection, and using only accuracy and time for parent selection), the system initially outpaces \method{} but quickly saturates, unable to push further without accumulated knowledge to guide its exploration. Without multi-agent feedback, a similar pattern emerges at a higher ceiling. It shows a rapid early gain followed by stagnation, as the system cannot diagnose \emph{why} architectures succeed or fail, and thus cannot refine its hypotheses.
Without hypothesis selection, the system shows steady but limited progress, as the memory accumulates evidence but cannot direct exploration toward the most informative experiments. Without parent selection (replaced with greedy, accuracy-based selection), the system follows a trajectory similar to that of the variant without hypothesis-driven search, confirming that intelligent parent selection is critical for escaping local optima. All four variants plateau well below \method{}'s 94.1\%. \method{}'s slower start reflects the cost of deliberately exploring uncertain hypotheses, a cost that pays off when it breaks through where the ablated variants cannot.

We also compare alternative parent selection strategies while holding all other components fixed (Fig.~\ref{fig:main_result}, right). Selection strategy used in previous works~\cite{si2026towards, novikov2025alphaevolvecodingagentscientific}, Exploration-Exploitation annealing (EE annealing) starts at 50\% exploration and anneals toward exploitation. It discovers high-accuracy architectures faster than other methods but saturates shortly after, as its fixed schedule cannot adapt to accumulated knowledge. Greedy selection plateaus earliest and lowest, exhibiting the exploitation collapse reported in prior work~\cite{zhang2025darwingodelmachineopenended, agrawal2026gepareflectivepromptevolution}. DGM-style selection~\cite{zhang2025darwingodelmachineopenended}, which weights parents by fitness and a novelty bonus inversely proportional to offspring count, fares only slightly better. Notably, random selection outperforms both greedy and DGM-style methods, highlighting that when hypothesis memory and multi-agent feedback are present, even undirected exploration can be effective. \method{} is the only method that continues improving throughout the full search, because it grounds its exploration decisions in accumulated knowledge rather than following a fixed schedule or relying on fitness alone.

\subsection{Cross-Dataset Generalization}
\label{sec:cross_dataset_generalization}

\begin{table}[t]
    \centering
    \small
    \caption{
    We report results on CIFAR-100 and Tiny-ImageNet for GSTN which was discovered on CIFAR-10. Results from manually engineered models of similar and larger capacity are included for reference.}
   \label{tab:cross_dataset_generalization}
    \footnotesize
    \setlength{\tabcolsep}{4pt}
    \renewcommand{\arraystretch}{1.15}
    \begin{tabular}{l c cc cc cc}
        \toprule
        \textbf{Model} & \textbf{Params (M)} &
        \multicolumn{2}{c}{\textbf{CIFAR-10}} &
        \multicolumn{2}{c}{\textbf{CIFAR-100}} &
        \multicolumn{2}{c}{\textbf{Tiny-ImageNet}} \\
        \cmidrule(lr){3-4}\cmidrule(lr){5-6}\cmidrule(lr){7-8}
        & & Acc@1 & Acc@5 & Acc@1 & Acc@5 & Acc@1 & Acc@5 \\
        \midrule
        ResNet-18     & 11.7 & 95.4 & 99.8 & 78.5 & 94.1 & 69.3 & 84.8 \\
        \midrule 
        MobileNet V3  & 2.5  & 95.5 & 99.9 & 73.0 & 92.5 & 58.5 & 83.4 \\
        ShuffleNet V2 & 1.4  & 90.1 & 99.4 & 67.5 & 85.2 & 50.9 & 73.4 \\
        SqueezeNet    & 1.2  & 91.1 & 99.7 & 67.3 & 83.8 & 54.7 & 77.1 \\
        \midrule
        GSTN (ours) & 0.9 & 94.1 & 99.6 & 72.6 & 91.7 & 58.1 & 81.7 \\
        \bottomrule
    \end{tabular}
\end{table}

Tab.~\ref{tab:cross_dataset_generalization} shows that the architecture discovered on CIFAR-10 transfers well to harder datasets. With only parameters, \textbf{GSTM} achieves \textbf{72.6/91.7} on CIFAR-100 and \textbf{58.1/81.7} on Tiny-ImageNet, matching MobileNetV3\textbf{0.9M}\begin{wraptable}{r}{0.56\linewidth} 
\centering
\footnotesize
\setlength{\tabcolsep}{1.3pt}
\renewcommand{\arraystretch}{1.18}
\caption{\textbf{Domain-specific architecture discovery on MedMNIST classification.}
We report top-1 accuracy (\%) on DermalMNIST, TissueMNIST, and BreastMNIST. ``--'' indicates the result is not reported. Best in \textbf{bold} and second-best \underline{underlined}.}
\label{tab:domain_specific_arch_discovery}
\begin{tabular}{@{}lccc@{}}
\toprule
\textbf{Method}
& \makecell{\scriptsize\textbf{Dermal}\\[-2pt]\scriptsize\textbf{MNIST}}
& \makecell{\scriptsize\textbf{Tissue}\\[-2pt]\scriptsize\textbf{MNIST}}
& \makecell{\scriptsize\textbf{Breast}\\[-2pt]\scriptsize\textbf{MNIST}} \\
\midrule
ResNet-18~\cite{he2016deep,manzari2025medical}            & 75.4 & 68.1 & 83.3 \\
ResNet-50~\cite{he2016deep,manzari2025medical}      & 73.1 & 68.3 & 82.8 \\
ViT~\cite{dosovitskiy2021an, chowdary2024med}     & 73.9 & --   & --   \\
Swin~\cite{liu2021swin,chowdary2024med} & 75.3 & -- & -- \\
\midrule
MedViTV1-L~\cite{manzari2023medvit} & 77.3 & 68.3 & 88.5 \\
MedMamba-B~\cite{yue2024medmamba}   & 75.7 & --   & 89.1 \\
MedFormer~\cite{chowdary2024med}    & 78.3 & --   & --   \\
NQNN~\cite{rahman2025nqnn}          & 80.4 & --   & --   \\
Med-LEGO~\cite{zhu2025med}          & 73.9 & --   & --   \\
PRADA~\cite{jang2025prada}          & 81.3 & --   & --   \\
MedNNS~\cite{mecharbat2025mednns}   & 79.7 & 69.2 & \textbf{92.3} \\
\underline{MedViTV2-L~\cite{manzari2025medical}} & \underline{81.7} & \underline{71.6} & 91.0 \\
\midrule
\textbf{Ours}                      & \textbf{82.1} & \textbf{73.9} & \underline{91.7} \\
\bottomrule
\vspace{-0.2in}
\end{tabular}

\vspace{-0.8\baselineskip}
\end{wraptable} within \textbf{0.4} Top-1 on both while using $\sim$2.8$\times$ fewer parameters. It also outperforms similarly lightweight baselines such as ShuffleNetV2 and SqueezeNet by 5--7 Top-1 points, suggesting that the discovered mechanisms transfer beyond CIFAR-10. Although ResNet-18 remains stronger in absolute accuracy, it requires roughly 13$\times$ more parameters, leaving GSTM on a favorable accuracy--efficiency frontier.

\vspace{-0.05in}
\subsection{Domain-Specific Architecture Discovery}
To demonstrate that \method{} can be applied beyond general visual recognition, we conduct an independent discovery run on DermalMNIST and evaluate the discovered architecture on three MedMNIST~\cite{yang2023medmnist} tasks (Tab.~\ref{tab:domain_specific_arch_discovery}). Our discovered architecture achieves the strongest performance on DermalMNIST (82.1\%) and TissueMNIST (73.9\%), outperforming the best reported results by +0.4\% and +2.3\%, respectively. Notably, these gains are obtained without initializing from a predefined seed backbone, but it is designed specifically for the medical domain. It further showcases that \method{} can favor downstream applications.
On BreastMNIST, our method remains competitive, achieving 91.7\% accuracy, only 0.6\% below the best-performing specialized baseline. Overall, these results suggest that active hypothesis exploration and memory-guided selection can discover effective architectures even for medical imaging domains within a computation budget.

\newpage

\section{Analysis}
\label{sec:analysis}

\subsection{Discovered Architectures}
We introduce three representative architectures discovered by \method{} that achieved the highest classification accuracy. Each uses a structurally distinct approach to efficiently aggregate global context without quadratic attention.      

\vspace{-1em}
\begin{itemize}
    \item \textbf{GSTN (94.11\%).} This design augments a lightweight three-stage ResNet backbone with a Global Shape Token (GST) module that introduces a small bank of learned global vectors as intermediaries for sparse global routing. Spatial features are softly assigned to these vectors via cosine similarity, and the aggregated global signal is residually blended back, providing content-adaptive global context 
    without attention. Learned global tokens are conceptually related to inducing points~\cite{lee2019set} and register tokens~\cite{darcet2024vision}.

    \item \textbf{Hierarchical Hub Routing Network (93.57\%).} HHRN replaces dense attention with a hub-mediated sparse routing mechanism. Tokens are softly assigned to a small set of learned hub vectors via cosine similarity. Hubs aggregate token messages, exchange information among themselves through a sparse GNN with top-$k$ adjacency, and broadcast refined corrections back to tokens. 
    The token-to-hub routing is structurally related to Slot Attention~\cite{locatello2020object}, while the hub-to-hub GNN shares design principles with Vision GNN~\cite{han2022vision}.

    \item \textbf{Band-Aware Wavelet Token Mixer (91.22\%).} BA-WTM+ performs explicit frequency-domain decomposition by splitting features into low-frequency and high-frequency channel halves via a learned analysis transform. A FiLM controller~\cite{perez2018film} conditioned on the low-frequency stream modulates the high-frequency bands, and a sigmoid-gated cross-band residual lets low-pass shape priors suppress spurious high-frequency textures. The entire pipeline uses only efficient $1 \times 1$ and depthwise convolutions without any attention mechanism. The band-split design relates to Octave Convolution~\cite{chen2019drop} and WaveMLP~\cite{tang2022image}.
  
\end{itemize}

\subsection{Discovered Hypotheses}
\method{} hypothesis memory accumulated 117 hypotheses throughout the discovery, of which 19 reached confirmed status (confidence > 0.7), 95 remained uncertain and subject to ongoing refinement, and only 3 were actively refuted. Among the confirmed hypotheses, three findings stand out. hyp\_44 (confidence 0.601, 3 supporting / 2 contradicting), which directly informed our best-performing architecture, suggests that adding as few as 3 to 4 small learnable global tokens that collect spatial summaries from all tokens and broadcast shape-aware updates back is sufficient to break texture dominance, a surprisingly economical intervention with outsized effect.  
hyp\_17 (confidence 0.830, 3 supporting / 0 contradicting) suggests that separating shape-oriented and texture-oriented token channels into dedicated representation banks, each compressed at its natural rate, recovers the ability of the network to attend to global object outlines in shape-reliant classes, though further validation is warranted. hyp\_26 (confidence 0.717, 3 supporting / 0 contradicting) offers a promising direction where introducing small learnable positional offsets into the feature transform pipeline gives the network a subtle sense of spatial position, potentially enabling it to distinguish asymmetric or location-dependent patterns at negligible cost. These findings suggest \method{} is not merely finding better architectures, but learning why certain designs succeed.

\begin{figure*}[t]
    \centering
    \includegraphics[width=0.38\textwidth]{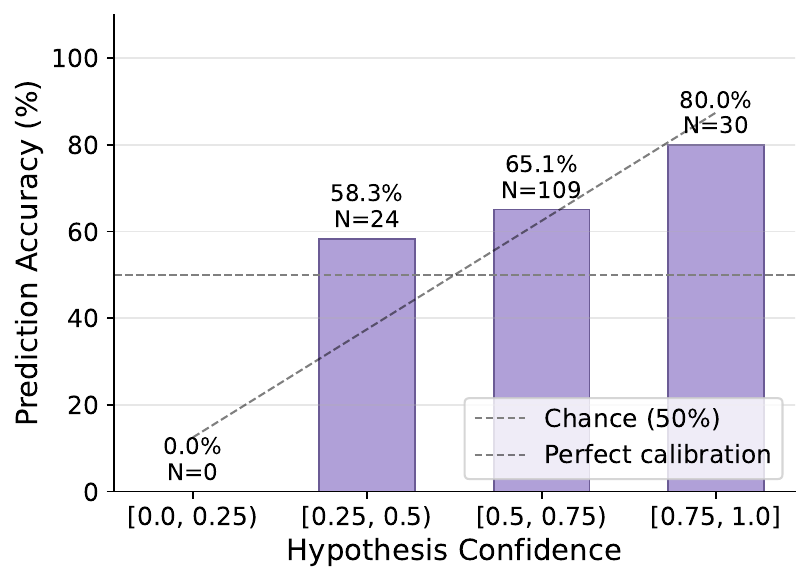}
    \hfill
    \includegraphics[width=0.58\textwidth]{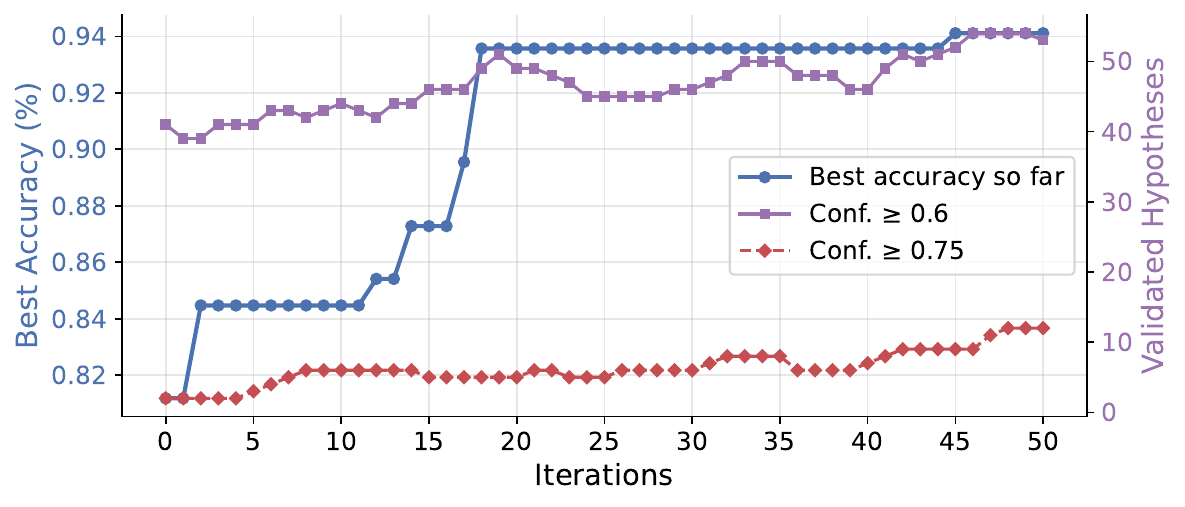}
    \caption{\textbf{Analysis of hypothesis memory over 50 iterations on CIFAR-10.} \textbf{Left: Hypothesis prediction accuracy.} Higher-confidence hypotheses predict experimental outcomes more accurately. \textbf{Right: Knowledge accumulation curve.} Best accuracy achieved and validated hypothesis count grows together.}
    \label{fig:hypo_analysis1}
\end{figure*}

\para{Hypothesis Prediction Accuracy.}
To evaluate whether the system confidence scores carry a meaningful predictive signal, we measure how often a hypothesis correctly predicts the direction of accuracy change when applied in an experiment.
Each hypothesis predicts that a specific architectural choice will have a positive or negative effect on performance. When the hypothesis is selected, we compare the resulting architecture accuracy to its parent node: the prediction is correct if the child improves over the parent when a positive effect was predicted, or degrades when a negative effect was predicted.
We bin all hypothesis-experiment pairs by the hypothesis's confidence at the time of testing. As shown in Figure~\ref{fig:hypo_analysis1} (Left), prediction accuracy increases monotonically with confidence: hypotheses in the [0.25, 0.5) bin predict correctly 58\% of the time (N=24), rising to 65\% in [0.5, 0.75) (N=109) and 80\% in [0.75, 1.0] (N=30). All bins exceed the 50\% chance baseline, and the lowest empty bin indicates that the system does not retain hypotheses lacking supporting evidence. The monotonic trend demonstrates that the confidence update mechanism produces scores that are meaningfully calibrated to actual predictive accuracy, not merely artifacts of the update rule.

\para{Knowledge Accumulation over Generation.}
We examine whether the accumulation of validated hypotheses correlates with improvements in the best architectures discovered. Figure~\ref{fig:hypo_analysis1} (Right) plots the best accuracy found so far alongside the count of validated hypotheses at two confidence thresholds (0.6 and 0.75) over the chronological sequence of experiments. 
The two curves show clear co-movement: the major accuracy jumps between nodes 15 and 25, where the best accuracy rises from 85\% to 93.5\%, coincide with a rapid increase in validated hypotheses at the 0.6 threshold. After node 30, both curves plateau together. Then there is a slight increase around node 45, followed by a final accuracy increase. This U-shaped correlation pattern is consistent with an explore-then-exploit dynamic: early generations build foundational knowledge, mid-generations diversify (temporarily weakening the correlation), and late generations consolidate validated knowledge into top-performing architectures. This pattern suggests that the system's performance gains are associated with the growth of its validated knowledge base rather than random exploration, and that the rate of discovery slows as the hypothesis space becomes increasingly explored.

\para{Cross-Lineage Knowledge Transfer.}\\[-0.5em]
\noindent\begin{minipage}[t]{0.69\linewidth}
\vspace{0pt}
A key question for any knowledge accumulation system is whether its learned principles generalize beyond the context in which they were discovered. \method{} maintains five independent evolutionary lineages originating from different root architectures. We classify each hypothesis application as within-lineage (hypothesis originated from the same root lineage) or cross-lineage (hypothesis originated from a different lineage), and measure whether the application led to an accuracy improvement. As shown in Figure~\ref{fig:cross_lineage}, cross-lineage applications succeed at 65\% (N=171), comparable to within-lineage success at 57\% (N=93). Notably, the system applies hypotheses across lineages nearly twice as often as within lineages, indicating the selection mechanism actively shares knowledge across independent branches. The comparable success rates demonstrate that the hypotheses capture transferable design principles rather than lineage-specific artifacts.
\end{minipage}
\hfill
\begin{minipage}[t]{0.27\linewidth}
\vspace{0pt}
\centering
\includegraphics[width=\linewidth]{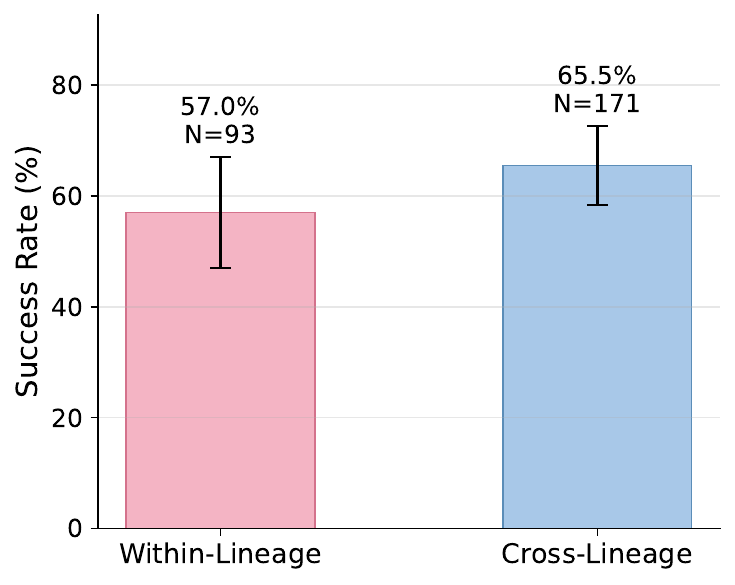}
\captionof{figure}{\footnotesize Cross-lineage hypothesis applications succeed at a comparable rate to within-lineage ones, indicating transferable design principles.}
\label{fig:cross_lineage}
\end{minipage}

\section{Conclusion}
\label{sec:conclusion}
We introduced \method{}, a multi-agent framework that reframes automated neural architecture discovery as hypothesis-driven scientific inquiry. By maintaining a trajectory tree and a hypothesis memory bank, \method{} separates where to search from what to test, and answers both using accumulated empirical evidence rather than undirected trial and error. This yields GSTN, a 0.9M parameter architecture reaching 94.11\% on CIFAR-10 that transfers competitively to CIFAR-100 and Tiny-ImageNet, while accumulating transferable design knowledge whose predictive accuracy grows with confidence. We believe the core insight, that autonomous systems should reason explicitly about what is known and what remains uncertain, extends well beyond architecture search toward machine-driven scientific discovery more broadly.

\bibliographystyle{plain}
\bibliography{main}

\appendix
\newpage
\section{Experiment Details}
In this section, we provide implementation details for reproducing our experiments.
All discovery runs use the same multi-agent pipeline described in Section 3 of the main paper and only the dataset-specific training recipe differ between CIFAR-10 and MedMNIST.

\subsection{Experiment Setups}

All agents in the pipeline use GPT-5-mini with a maximum output length of 32\,768 tokens.
Generation agents (Idea, Architect, Coding, and Feedback) use temperature 0.7, synthesis and memory agents use 0.3, and the redundancy-filter LLM Judge uses 0.1. For the redundancy filtering agent, we use Gemini Embedding API (\texttt{gemini-embedding-001}, 256-dim) and the top-3 most similar archived concepts are retrieved by cosine similarity.
All experiments are executed on a single NVIDIA A40 GPU.
Each architecture is trained with a hard wall-clock timeout of 30 minutes.

\subsection{CIFAR-10 Training Protocol}
\label{sec:supp_cifar}
Table~\ref{tab:cifar_train} shows the training protocol of CIFAR-10.

\begin{table}[h]
\centering
\caption{Default CIFAR-10 training configuration. All values may be overridden by the Coding Agent via the generated \texttt{config.py}. Additionally, new hyperparameters can be added by the Coding Agent.}
\label{tab:cifar_train}
\small
\begin{tabularx}{\linewidth}{@{}lX@{}}
\toprule
\textbf{Setting} & \textbf{Value} \\
\midrule
Resolution        & $32 \times 32$ (native) \\
Normalisation     & $\mu=(0.4914, 0.4822, 0.4465)$, $\sigma=(0.2470, 0.2435, 0.2616)$ \\
Augmentation      & Random horizontal flip (50\%), random translation ($\pm 4$\,px, reflect pad), Cutout ($12 \times 12$) \\
Data loader       & GPU-resident (entire dataset in VRAM); all augmentation on-GPU; FP16, channels-last \\
\midrule
Optimiser         & SGD (lr$_{\text{default}}$\,=\,0.1, momentum\,=\,0.9, weight decay\,=\,$5{\times}10^{-4}$, Nesterov) \\
LR schedule       & Cosine annealing with linear warmup (5 epochs) \\
Label smoothing   & $\epsilon = 0.1$ (default) \\
Gradient clipping & $\lVert g \rVert \leq 1.0$ \\
Precision         & FP16 mixed-precision \\
Batch size        & 1\,024 (default) \\
\midrule
Wall-clock budget & 1\,800\,s (30\,min) per experiment \\
Error recovery    & Up to 10 retries on code errors \\
HP refinement     & Up to 5 steps after first successful run \\
\bottomrule
\end{tabularx}
\end{table}

\newpage
\subsection{MedMNIST Training Protocol}
\label{sec:supp_medmnist}
Table~\ref{tab:medmnist_train} shows the training protocol of MedMNIST.
\begin{table}[h]
\centering
\caption{Default MedMNIST training configuration. All values may be overridden by the Coding Agent via the generated \texttt{config.py}. Additionally, new hyperparameters can be added by the Coding Agent. }
\label{tab:medmnist_train}
\small
\begin{tabularx}{\linewidth}{@{}p{0.28\linewidth}X@{}}
\toprule
\textbf{Setting} & \textbf{Value} \\
\midrule
Datasets           & DermaMNIST (7 cls) \\
Resolution         & $224 \times 224$ (aligned with the MedViTV2 SOTA protocol) \\
Channel conversion & Grayscale $\to$ RGB (\texttt{as\_rgb=True}) \\
Normalisation      & $\mu=(0.5, 0.5, 0.5)$, $\sigma=(0.5, 0.5, 0.5)$ \\
Augmentation       & \texttt{RandomResizedCrop(224)}, \texttt{AugMix} (sev.=3, width=3, $\alpha=0.4$), \texttt{RandomHorizontalFlip} ($p=0.4$) \\
Test/val transform & \texttt{Resize(224)} + normalisation only \\
\midrule
Optimiser          & Same SGD defaults as CIFAR-10 (\Cref{tab:cifar_train}) \\
Precision          & FP16 mixed-precision \\
Batch size         & 128 (4 data-loader workers, pinned memory) \\
\midrule
Wall-clock budget  & 1\,800\,s (30\,min) per experiment \\
Error recovery     & Same as CIFAR-10 (up to 10 retries, 5 refinement steps) \\
\bottomrule
\end{tabularx}
\end{table}

\section{Implementation Details}
\label{sec:impl_details}

This section specifies the exact input/output contracts and behavioral modes of each component in the \method{} pipeline, complementing the high-level description in Section 3.

\subsection{Idea Agent} 
The Idea Agent operates in two distinct modes with different input/output contracts.
Here we first specify the root node's input/output and behavior:

\begin{agentinput}
Research direction and a curated set of literature references relevant to the target domain.
\end{agentinput}

\begin{agentoutput}
$K$ diverse architecture proposals generated in a \emph{single} batch LLM call. Each proposal contains:\\[2pt]
\textbullet~\texttt{title}, \texttt{description}, \texttt{intuition}, \texttt{novelty} claim\\
\textbullet~\texttt{architecture\_spec}: \texttt{core\_ideas}, \texttt{core\_blocks} (named modules with descriptions), \texttt{network\_structure} (block connectivity), \texttt{tunable\_aspects}, \texttt{invariants} (design principles to preserve)\\
\textbullet~Up to $K_{\mathrm{hyp}}$ new hypotheses in structured format:\\[2pt]
\texttt{\small IF [architectural choice] IN [scope], THEN [predicted effect], BECAUSE [mechanism]. DISPROVED IF [falsification criterion].}\\[2pt]
Each hypothesis additionally specifies \texttt{scope}, \texttt{prediction}, \texttt{falsification\_criteria}, \texttt{tags}, and \texttt{initial\_confidence}~$= 0.5$.
\end{agentoutput}

\begin{agentbehavior}
Generates $K$ fundamentally distinct architectures from scratch in one LLM call. Batch generation ensures inter-idea diversity (all $K$ proposals in one call rather than $K$ independent calls). Guided by curated literature references. No parent context or hypothesis memory available yet.
\end{agentbehavior}

\noindent Here is the behavior of the Idea Agent in the evolution mode (generation $\geq 1$):
\begin{agentinput}
Five context blocks injected as template variables into the prompt:\\[2pt]
1.~\texttt{\{research\_direction\}}: the research direction string\\
2.~\texttt{\{parent\_architecture\}}: parent node's full brainstorming output (title, description, intuition, novelty, architecture spec) + performance metrics (accuracy, training time, novelty score)\\
3.~\texttt{\{feedback\_summary\}}: concatenated outputs from all feedback agents, including per-agent reasoning, actionable recommendations, hypothesis updates with current confidence from $\mathcal{M}_t$, and newly proposed hypotheses\\
4.~\texttt{\{hypothesis\_memory\}}: compiled hypothesis context from $\mathcal{M}_t$, grouped into confirmed ($c > 0.75$), refuted ($c < 0.25$), and uncertain patterns, each with full evidence log and agent attribution\\
5.~Selected hypothesis $h^\star$ to test
\end{agentinput}

\begin{agentoutput}
A single evolved architecture containing:\\[2pt]
\textbullet~Structured \texttt{reasoning} trace: \texttt{parent\_analysis} (what worked), \texttt{failure\_analysis} (what failed), \texttt{hypothesis\_usage} (which hypotheses guide design), \texttt{proposed\_changes} (targeted modifications)\\
\textbullet~Architecture fields: \texttt{title}, \texttt{description}, \texttt{architecture\_spec}, etc.\\
\textbullet~\texttt{existing\_hypotheses}: referenced hypothesis IDs (e.g., \texttt{["hyp\_3", "hyp\_7"]})\\
\textbullet~Up to $K_{\mathrm{hyp}}$ new hypotheses motivated by the proposed design
\end{agentoutput}

\begin{agentbehavior}
Hypothesis-guided design conditioned on parent context. Makes targeted 1--2 aspect changes (not full redesigns). Builds on confirmed patterns, investigates uncertain hypotheses, avoids refuted patterns.
\end{agentbehavior}

\subsection{Coding Agent}
The Coding Agent operates in three modes, each with distinct input/output.

\noindent First, here is the specific details for the initial generation (iteration~1):

\begin{agentinput}
Architecture specification from the Idea Agent, problem specification (dataset, input shape, number of classes), and implementation notes from $\ell_t(h^\star)$ recording failure modes and recommended practices from prior experiments testing the same hypothesis.
\end{agentinput}

\begin{agentoutput}
Two files: \texttt{model.py} (PyTorch \texttt{nn.Module}) and \texttt{config.py} (hyperparameter dictionary).\\[2pt]
\emph{Constraints:} $<$10M parameters, mixed-precision (AMP) support, channels-last memory format, training must complete within $t_{\max}$.
\end{agentoutput}

\newpage
\noindent Here is the specific details for Error recovery (fix mode):

\begin{agentinput}
Previous \texttt{model.py}, error traceback, and the original architecture specification.
\end{agentinput}

\begin{agentoutput}
Corrected \texttt{model.py}; the existing \texttt{config.py} is reused.
\end{agentoutput}

\begin{agentbehavior}
Retries up to $R_{\max}$ times. Each retry receives the latest traceback, enabling iterative debugging.
\end{agentbehavior}

We specify the details for the Hyperparameter refinement (refine mode):

\begin{agentinput}
Previous \texttt{config.py}, training logs (loss curves, accuracy progression, convergence diagnostics), and the current \texttt{model.py}.
\end{agentinput}

\begin{agentoutput}
Updated \texttt{config.py}; the existing \texttt{model.py} is reused.
\end{agentoutput}

\begin{agentbehavior}
Runs for up to $F_{\max}$ steps with early stopping if accuracy plateaus between consecutive iterations.
\end{agentbehavior}
\subsection{Redundancy Filtering Agent}
Before execution, a two-stage filter prevents re-exploration of previously visited concepts.

\noindent\textbf{Stage 1: Embedding-based retrieval.}
The candidate architecture's concept description is embedded (256-dimensional, Gemini Embedding API) and the top-$k{=}3$ most similar archived concepts in $\mathcal{T}_t$ are retrieved by cosine similarity.

\noindent\textbf{Stage 2: LLM judge.}

\begin{agentinput}
Candidate architecture (title, description, generated code) together with the top-3 retrieved concepts from $\mathcal{T}_t$.
\end{agentinput}

\begin{agentoutput}
Structured judgment:\\[2pt]
\textbullet~\texttt{novel}: boolean decision\\
\textbullet~\texttt{reasoning}: explanation of the decision\\
\textbullet~\texttt{most\_similar\_to}: ID of the most similar existing concept\\
\textbullet~\texttt{shared\_principles}: what the candidate shares with existing work\\
\textbullet~\texttt{new\_contribution}: what is genuinely new
\end{agentoutput}

\begin{agentbehavior}
Compares structural design principles rather than surface-level differences (e.g., different hyperparameters or activation choices do not constitute novelty). Same innovation claim + same structural pattern = duplicate. If rejected, the Idea Agent may regenerate with explicit instructions to avoid the flagged overlap, for up to 2 retries.
\end{agentbehavior}

\subsection{Feedback Agents}
After each experiment, four specialized agents analyze the outcome from complementary perspectives.
All agents share a common output schema containing
\texttt{reasoning}, \texttt{actionable\_feedback}, and
\texttt{hypothesis\_updates}. Each hypothesis update contains
\texttt{evidence\_type} $\in \{\texttt{supports}, \texttt{contradicts}, \texttt{neutral}\}$,
\texttt{strength} $\in [0,1]$, \texttt{reasoning}, and
\texttt{new\_hypotheses}.

\noindent We specify the details of the Quantitative Feedback Agent:

\begin{agentinput}
Training curves (loss and accuracy per epoch), final top-1/top-5 accuracy, per-class accuracy breakdown, parameter count, FLOPs, and training time.
\end{agentinput}

\begin{agentoutput}
Statistical analysis of convergence patterns, efficiency characterization, and hypothesis evidence. Identifies overfitting signatures, class-specific weaknesses, and computational bottlenecks.
\end{agentoutput}

\noindent We specify the details of the Qualitative Feedback Agent (VLM):

\begin{agentinput}
$N_{\mathrm{err}}$ highest-confidence misclassified test images with corresponding heatmap overlays. Heatmap method is auto-detected: GradCAM (CNN-dominant), Attention Rollout (transformer-dominant), Input Gradient Saliency (universal fallback, e.g.\ MLP-Mixers).
\end{agentinput}

\begin{agentoutput}
Per-image analysis of attention patterns, texture vs.\ shape bias identification, confusion patterns between visually similar classes, and hypothesis evidence about failure modes.
\end{agentoutput}

\noindent We specify the details of the Causal Feedback Agent: 

\begin{agentinput}
Parent and child architecture specifications, structural diff between them, and performance delta ($\Delta\text{accuracy}$, $\Delta\text{training time}$).
\end{agentinput}

\begin{agentoutput}
Causal attribution of observed performance changes to specific structural modifications (e.g., ``replacing batch normalization with layer normalization improved accuracy by 1.2\%''). Produces directed hypothesis updates linking architectural changes to outcomes.
\end{agentoutput}

\noindent We specify the details of the Diagnostic Feedback Agent (failure/timeout only): 

\begin{agentinput}
Error traceback or timeout logs, the generated \texttt{model.py}, and the architecture specification.
\end{agentinput}

\begin{agentoutput}
Root cause analysis (e.g., OOM, numerical instability, excessive FLOPs) and structured implementation notes:\\[2pt]
\textbullet~\texttt{common\_failures}: failure modes with descriptions, recorded into $\ell_t(h)$\\
\textbullet~\texttt{recommended\_practices}: concrete guidance to avoid recurrence\\[2pt]
These notes persist in the hypothesis memory and are provided to the Coding Agent in future iterations testing the same hypothesis, enabling the system to learn from implementation failures.
\end{agentoutput}

\subsection{Hypothesis Synthesis Agent}
\begin{agentinput}
Concatenated outputs of all feedback agents (hypothesis updates and new hypothesis proposals) and the current hypothesis memory $\mathcal{M}_t$.
\end{agentinput}

\begin{agentoutput}
Consolidated update containing:\\[2pt]
\textbullet~\texttt{hypothesis\_updates}: deduplicated evidence updates for existing hypotheses, each with \texttt{hyp\_id}, \texttt{evidence\_type}, \texttt{strength}, and cross-agent \texttt{reasoning}\\
\textbullet~At most $K_{\mathrm{synth}}{=}2$ \texttt{new\_hypotheses}, each in the structured IF/THEN/BECAUSE/DISPROVED~IF format with \texttt{scope}, \texttt{prediction}, \texttt{falsification\_criteria}, and \texttt{tags}\\
\textbullet~\texttt{implementation\_notes}: failure modes and recommended practices keyed to hypothesis IDs
\end{agentoutput}

\begin{agentbehavior}
All feedback agent outputs are processed in a single LLM call. Resolves three types of conflicts:\\[2pt]
(i)~overlapping hypothesis updates from different agents are merged, with synthesized reasoning citing all contributing perspectives;\\
(ii)~contradictory evidence assessments are resolved by weighing the specificity of each agent's analysis;\\
(iii)~redundant new hypothesis proposals are collapsed.
\end{agentbehavior}

\noindent\textbf{Quality gate.}
Every new hypothesis must pass a 7-dimension quality gate before admission to $\mathcal{M}_t$:
\label{sec:impl_synthesis}

\begin{tcolorbox}[colback=orange!5!white, colframe=orange!70!black, title=\textbf{7-Dimension Hypothesis Quality Gate}]
\begin{enumerate}[nosep,leftmargin=*]
    \item \emph{Mechanistic}---explains \emph{why} via a causal mechanism, not just correlation
    \item \emph{Scoped}---states where the hypothesis applies and where it does not
    \item \emph{Predictive}---specifies the expected direction and magnitude of effect
    \item \emph{Falsifiable}---defines what experimental result would disprove it
    \item \emph{Novel}---is not a restatement of an existing hypothesis in $\mathcal{M}_t$
    \item \emph{Transferable}---applies across architectures, not just the current model
    \item \emph{Actionable}---can be tested with a single targeted architectural change
\end{enumerate}
Hypotheses failing any dimension are either revised by the synthesis agent or discarded.
\end{tcolorbox}

\newpage
\subsection{Hypothesis Memory Bank}
The hypothesis memory bank $\mathcal{M}_t$ stores per-hypothesis records and provides structured retrieval for downstream agents.

\begin{tcolorbox}[colback=violet!5!white, colframe=violet!60!black, title=\textbf{Per-hypothesis storage.}]
Each hypothesis $h \in \mathcal{M}_t$ maintains:
\begin{itemize}[nosep,leftmargin=*]
    \item Confidence $c_t(h) \in [0.01, 0.99]$, initialized at $c_0 = 0.5$;
    \item An evidence log: list of $(\text{node\_id}, \text{type}, \text{strength}, \text{reasoning}, \text{agent})$ tuples
    \item Implementation notes $\ell_t(h)$: accumulated failure modes with frequency and recommended practices
    \item Bidirectional connections to related hypotheses
    \item Structured fields: \texttt{scope}, \texttt{prediction}, \texttt{falsification\_criteria}
    \item Provenance: \texttt{created\_by} agent identifier and source node.
\end{itemize}
\end{tcolorbox}

\noindent\textbf{Confidence update.}
Evidence updates follow Equation 1 from the main paper with learning rate $\eta = 0.20$.
Hypotheses are classified as \emph{confirmed} when $c > 0.75$, \emph{refuted} when $c < 0.25$, and \emph{uncertain} otherwise.


\subsection{Trajectory Tree Memory}

The trajectory tree $\mathcal{T}_t$ stores the complete research history as a forest of parent--child node relationships.

\begin{tcolorbox}[colback=teal!5!white, colframe=teal!60!black, title=\textbf{Per-node storage}]
Each \texttt{IdeaNode} $v \in \mathcal{T}_t$ stores:
\begin{itemize}[nosep,leftmargin=*]
    \item \texttt{idea}: the full brainstorming output dictionary;
    \item \texttt{architecture\_graph}: DAG representation (nodes and edges) for novelty scoring;
    \item \texttt{wl\_embedding}: 1024-dimensional Weisfeiler--Lehman embedding vector;
    \item \texttt{code\_attempts}: list of all \texttt{(model.py, config.py)} pairs attempted;
    \item \texttt{experiments}: training run results (accuracy, loss curves, timing, FLOPs);
    \item \texttt{feedback}: per-agent feedback records;
    \item Hypothesis tracking: \texttt{tested\_hypothesis\_id}, \texttt{hypothesis\_prediction}, \texttt{hypothesis\_evidence\_type}, and \texttt{hypothesis\_evidence\_strength};
    \item Scores: \texttt{best\_accuracy}, \texttt{novelty\_score}, \texttt{selection\_score}, \texttt{times\_selected}.
\end{itemize}
\end{tcolorbox}

\section{Additional Results}
\begin{figure}[thb]
  \centering
  \includegraphics[width=0.50\linewidth]{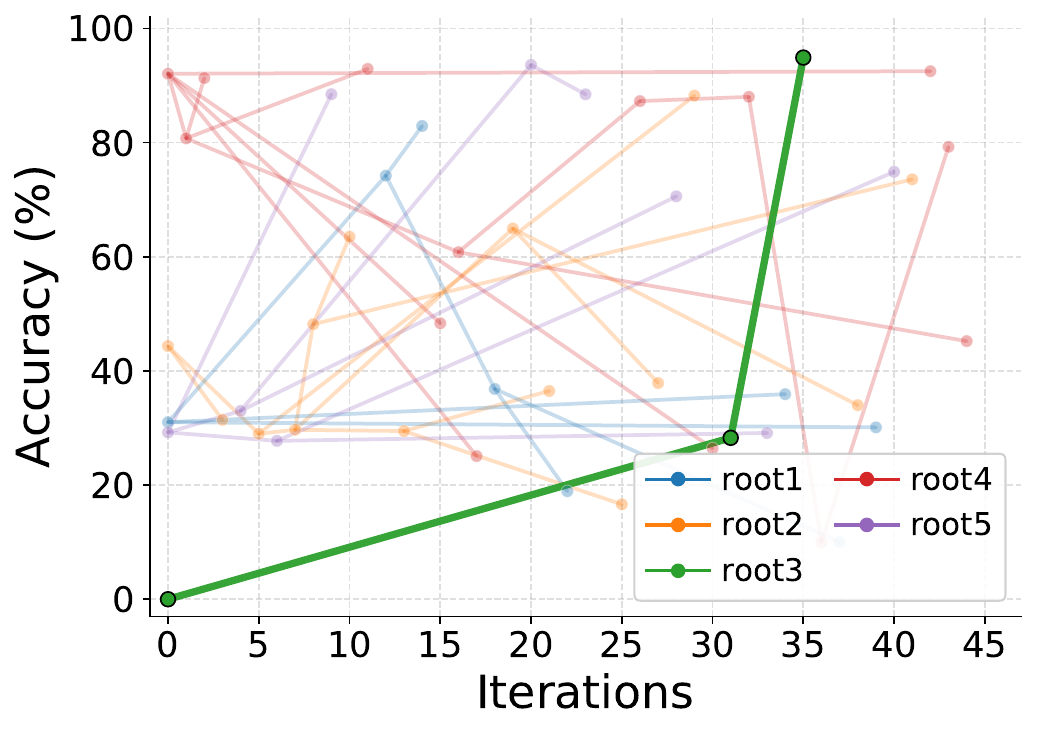}
  \caption{HyporExplore's run using Gemini-3.1-pro}
  \label{fig:gemini}
\end{figure}

Figure \ref{fig:gemini} visualizes \method{}'s run using Gemini 3.1 Pro across 45 iterations excluding the five root root nodes. The x-axis represents iteration order (0 for the 5 root ideas, then 1 through 45 for subsequent architectures), while the y-axis shows CIFAR-10 test accuracy (\%). 
Each color corresponds to one of the 5 root lineages: root1 (node\_0, "PolyMixer," 31.0\%), root2 (node\_1, "LatentMixer," 44.4\%), root3 (node\_2, "ScanMLP," 0.0\%), root4 (node\_3, "DilatedPatchMLP," 92.1\%), and root5 (node\_4, "HyperCubeMLP," 29.2\%). 
The thick green line highlights the best-performing lineage path, where it originates from root3's node\_2 ("ScanMLP"), which completely failed at 0.0\% accuracy due to repeated CUDA out-of-memory errors. Despite this unpromising start, \method{} selected root3 in Generation 31, due to its high exploration bonus from being unvisited and found a neural network with 28.3\% accuracy ("DualFreqScanMLP"). Then in Generation 35, it was selected and produced "PyramidGateMLP: Multi-Scale Pooled Gating with Full-Capacity Global Context", which achieved the highest accuracy of 94.9\%, a dramatic jump of +66.6 percentage points in a single generation and the overall best across all 50 evaluated architectures.

\section{Discovered Architectures}
This section presents the complete Idea Agent output and final implementation code for the three highest-performing architectures discovered.

\subsection{GST-Guarded NPIN (94.11\%)}
\label{sec:top3_node64}

\begin{ideabox}[title={Idea Agent Output: GST-Guarded NPIN: Sparse Global Shape Tokens + Per-Band Normalized Super-Particles}]

\textbf{Description.}
Build on NPIN-Guard (particle dynamics + EMA-stabilized super-particles + FiLM gating + band-split) and introduce two minimal, targeted changes:
(1)~add a very small set of stage-local Global Shape Tokens (GSTs, $G \leq 4$) that aggregate low-frequency token summaries via sparse top-$k$ routing and provide a compact, shape-focused global routing path;
(2)~apply per-band magnitude normalization to super-particle (slot) anchor messages before FiLM blending, plus make FiLM gate compute a GST-conditioned coherence score (GST-conditioned gate).
These changes are introduced to (a)~provide an explicit, tiny shape-oriented global aggregator that supplements semi-parametric super-particles and reduces reliance on many texture-driven slots, and (b)~prevent any anchor message (esp.\ high-frequency) from overwhelming token identity during broadcast.
All remaining NPIN-Guard components ($S{=}2$ local integrator, locality grid neighbor lookup, EMA+usage-normalized slot updates, band-split, per-band bookkeeping) are preserved so differences are causal and small.

Architecture pipeline:
PatchEmbed $\rightarrow$ BandSplit (DWConv low-pass, residual high-pass) $\rightarrow$ ParticleDynamicsBlock (per-band local neighbor gather; $S{=}2$ integrator; depthwise MLP PotentialNet) $\rightarrow$ SuperParticleCoarsen (per-band $K_b$ slots, soft assignment, EMA + per-slot usage normalization) $\rightarrow$ GST Module ($G$ small learned tokens per stage that collect low-frequency token summaries via sparse top-$k$ routing and compute compact GST states) $\rightarrow$ GST-conditioned FiLM Gate \& Per-band Anchor Normalization $\rightarrow$ FiLM-Gated Broadcast (anchor messages modulate tokens only when gated) $\rightarrow$ Token MLP \& Norm.

\textbf{Intuition.}
NPIN-Guard already demonstrated that stabilizing super-particle prototypes and FiLM-gating multiscale messages dramatically reduces texture-driven failures.
Two residual failure modes remain plausible:
(a)~some global shape groupings require a compact cross-token aggregator not easily captured by semi-static slots because slots are optimized for multi-band prototype coverage rather than explicitly coding global shape templates, and
(b)~large anchor messages (especially from high-frequency slots) can still overwhelm token identity despite gating if magnitudes vary across bands.
Adding a tiny set of learned GSTs supplies an explicit, low-capacity global routing channel that focuses on low-frequency/shape cues ($G$ small keeps compute trivial).
Per-band L2-normalization of anchor messages prevents magnitude-driven takeovers and makes blending with residual token features numerically stable, further reducing prototype monopolies.

\textbf{Novelty.}
The architecture fuses three mechanisms in a unique, minimal way:
(1)~particle-ODE local dynamics for efficient local mixing,
(2)~semi-parametric EMA-stabilized super-particles for multiscale aggregation, and
(3)~a tiny stage-local GST pathway that performs sparse top-$k$ low-frequency routing to provide explicit, compact shape aggregation.
The GSTs do not implement full cross-attention and are constrained ($G \leq 4$, sparse routing) to remain structurally distinct from standard global attention methods.
Per-band L2-normalization of anchor messages (applied before FiLM blending) combined with GST-conditioned gating is a novel practical stabilizer that directly targets prototype takeover while preserving local dynamics.

\textbf{Target Improvement:} both (accuracy and efficiency).

\textbf{Architecture Specification:}

\emph{Core Ideas:}
1)~Preserve NPIN particle dynamics (local neighbor ODE steps) for efficient local mixing.
2)~Keep EMA-stabilized, per-band super-particle slots with per-slot usage normalization to avoid drift.
3)~Add stage-local Global Shape Tokens (GSTs, $G\leq4$) that aggregate low-frequency token summaries via sparse top-$k$ routing ($O(G \cdot N)$ cheap ops) and broadcast compact shape updates back to tokens.
4)~Apply per-band L2-normalization to anchor messages before FiLM blending and compute FiLM gate as a GST-conditioned coherence score, so broadcasts occur only when GSTs and low-frequency tokens agree.

\emph{Core Blocks:}
\begin{itemize}[nosep,leftmargin=*]
    \item \textbf{PatchEmbed}: $3{\times}3$ conv stride~2 (or stride~1 + small pooling) $\rightarrow$ produce token features and initial learned-position offsets. Use small base channels ($C{=}64$) to control params.
    \item \textbf{BandSplit}: DWConv low-pass ($k{=}3$) $\rightarrow$ low-frequency stream; high-frequency $=$ residual (feat $-$ lowpass). Reduce per-band channels with grouped $1{\times}1$ conv to keep budget low.
    \item \textbf{ParticleDynamicsBlock}: For each band share neighbor grid; for each token gather neighbors within radius~$r$ (small). PotentialNet: depthwise MLP producing neighbor messages aggregated via sum/mean. Integrator: $S{=}2$ explicit steps updating per-token feature residual and small learned position offsets. Residual connections preserved.
    \item \textbf{SuperParticleCoarsen} (per-band): Soft assignment (temperature~$\tau$) of tokens to $K_b$ slots; aggregate normalized per-slot summaries; update prototypes via EMA: $\mathrm{proto} \leftarrow m \cdot \mathrm{proto} + (1{-}m)\cdot(\mathrm{agg}/(\mathrm{usage}+\varepsilon))$. Keep usage as EMA of assignment mass. Small learned gradient step permitted but heavily scaled.
    \item \textbf{GST Module} (stage-local): $G$ learnable tokens per stage ($G{=}1\ldots4$). Each token computes a tiny low-frequency key (MLP~$\rightarrow k_\mathrm{dim}$). Compute top-1 or top-2 scores for~$G$ (cheap when $G$ small). Aggregate token low-frequency summaries into chosen GSTs (sum or weighted sum). Update GST vectors with small gradient steps; optionally warmup EMA to stabilize early training.
    \item \textbf{GST-conditioned FiLM Gate \& Anchor Norm}: Per-band anchor vectors (super-particle~$\rightarrow$~token messages) L2-normalized independently. Compute gate $g_\mathrm{token} = \sigma(w \cdot \mathrm{cos\_sim}(\mathrm{token\_lf}, \mathrm{GST\_state}) + b)$. Gates multiplicatively scale FiLM parameters and broadcast mass.
    \item \textbf{FiLM-Gated Broadcast}: For each token, compute slot-weighted (soft assignment) anchor message but multiply by $g_\mathrm{token}$ and apply FiLM: $\mathrm{feat}' = \mathrm{feat} \cdot (1 + g_\mathrm{token} \cdot \gamma_\mathrm{slot}) + g_\mathrm{token} \cdot \beta_\mathrm{slot}$. Because anchors are normalized, the FiLM scale remains bounded.
    \item \textbf{Token MLP \& Norm}: Depthwise MLP + residual normalization to finalize block output.
\end{itemize}

\emph{Network Structure:}
3-stage backbone: Early stage (no coarsen): 3 ParticleDynamicsBlocks, base channels $C{=}48\!\rightarrow\!64$. Mid stage (first coarsen): 3 ParticleDynamicsBlocks with per-band slots ($K_\mathrm{low\_mid}{=}24$, $K_\mathrm{high\_mid}{=}12$) and GSTs $G_\mathrm{mid}{=}2$. Final stage: 2 ParticleDynamicsBlocks with coarsen ($K_\mathrm{low\_final}{=}8$, $K_\mathrm{high\_final}{=}4$) and GSTs $G_\mathrm{final}{=}1$. Global avg pooling $\rightarrow$ classifier head. Residuals across blocks; per-band bookkeeping passed forward. Use grouped $1{\times}1$ convs and depthwise ops to keep params low. Estimated param/FLOP budget: ${\sim}1.0$--$3.5$M params (config-dependent) and $<$200M FLOPs for CIFAR-size inputs.

\emph{Tunable Aspects:}
\begin{itemize}[nosep,leftmargin=*]
    \item GST count $G$ per stage (1--4)
    \item Super-particle counts $K_b$ per band/stage
    \item EMA momentum for prototypes (0.98--0.999)
    \item Assignment temperature $\tau$
    \item Neighbor radius $r$ and integrator steps $S$ ($S$ default $=2$)
    \item Anchor L2-norm scaling factor (post-normalization scale)
    \item Gate computation (cos-sim vs.\ MLP), thresholding
\end{itemize}

\emph{Invariants:}
\begin{itemize}[nosep,leftmargin=*]
    \item All local propagation inside ParticleDynamicsBlock uses only local neighbors (no all-to-all $N^2$ ops).
    \item Super-particles are semi-parametric EMA-updated slots with per-slot usage normalization retained.
    \item Bandwise bookkeeping (low/high) is preserved end-to-end and normalization is applied per-band, not globally.
    \item GSTs remain small ($G\leq4$) and perform sparse $O(G \cdot N)$ routing, not full $N{\times}N$ attention.
\end{itemize}

\textbf{New Hypotheses:}
\begin{enumerate}[nosep,leftmargin=*]
    \item \textbf{IF} a small number ($G \leq 4$) of stage-local Global Shape Tokens (GSTs) are added that aggregate low-frequency token summaries via sparse top-$k$ routing and broadcast compact shape-aware updates back to tokens \textbf{IN} NPIN-Guard-style particle+slot backbones on CIFAR-scale tasks, \textbf{THEN} overall shape-sensitive accuracy will increase and texture-driven confusions will decrease, \textbf{BECAUSE} GSTs provide a low-cost, explicit global routing channel specialized for low-frequency/shape cues that assembles spatially distributed evidence without relying on many texture-prone super-particle slots. \textbf{DISPROVED IF} adding GSTs (with controlled extra params and matched FLOPs) fails to reduce texture-driven misclassifications or GST activations remain near-zero across validation images.\\
    \emph{Tags:} \texttt{global-routing}, \texttt{sparse-routing}, \texttt{shape-aggregation}.
    \emph{Initial confidence:} 0.55.
    \emph{Connected:} \texttt{hyp\_44}, \texttt{hyp\_27}.\\
    \emph{Reasoning:} GSTs are a compact, low-capacity mechanism targeted to the same failure mode (shape starvation) identified previously. \texttt{hyp\_44} already suggested GSTs with moderate evidence; here we test a GST design constrained to sparse top-$k$ routing and low-frequency conditioning to keep overhead tiny and maximize shape focus.

    \item \textbf{IF} super-particle anchor messages are L2-normalized per-band before FiLM blending \textbf{IN} multi-band NPIN-style backbones, \textbf{THEN} the incidence of anchor-driven prototype-takeover misclassifications will decrease and shape-sensitive per-class recall will improve, \textbf{BECAUSE} bounding anchor magnitudes prevents high-frequency anchors from numerically overwhelming token identity vectors and allows gating/FiLM modulation to effectively preserve local shape evidence. \textbf{DISPROVED IF} per-band L2-normalization yields no measurable reduction in concentrated confusion mass onto dominant texture classes and no improvement in shape-sensitive per-class recall beyond random run variance.\\
    \emph{Tags:} \texttt{normalization}, \texttt{stability}, \texttt{anchor-magnitude}.
    \emph{Initial confidence:} 0.6.
    \emph{Connected:} \texttt{hyp\_58}, \texttt{hyp\_18}.\\
    \emph{Reasoning:} \texttt{hyp\_58} articulated the same mechanism; NPIN-Guard improvements hinted that message magnitude could be a remaining knob. Per-band normalization is simple, inexpensive, and falsifiable by measuring confusion mass and per-prototype assignment distributions.
\end{enumerate}

\textbf{Reasoning:}

\emph{Parent Analysis:}
NPIN-Guard preserved NPIN's core strength---local particle dynamics (learned potential + short ODE integration) with an efficient locality grid---for cheap and effective local mixing. The introduction of EMA-stabilized super-particles, per-slot usage-normalization, FiLM gating conditioned on low-frequency features, and band-splitting all correlated with a large, robust improvement in Top-1 accuracy (from ${\sim}19\%$ to ${\sim}57\%$), stable training dynamics, and reduced texture-dominated collapse.

\emph{Failure Analysis:}
Although NPIN-Guard made big gains, residual risks remain:
(1)~some global shape groupings may not be captured by slot prototypes alone, causing remaining confusions;
(2)~anchor messages, especially from high-frequency slots, can still have magnitude imbalances that numerically overwhelm token identity even when gated;
(3)~causal attribution was confounded because multiple interventions were introduced simultaneously in NPIN-Guard.

\emph{Hypothesis Usage:}
We prioritize \texttt{hyp\_13} (prototype drift mitigated by EMA + usage-normalization) and \texttt{hyp\_27} (FiLM-gated retrieval conditioned on low-frequency shape cues) as confirmed drivers---so we keep EMA + per-slot usage-normalization and low-frequency FiLM gating unchanged.
\texttt{hyp\_44} (GSTs) performed moderately in memory and directly motivates adding a tiny explicit global routing channel; we adopt this but constrain GSTs to be sparse and low-capacity to avoid large compute/overfitting.
\texttt{hyp\_18} and \texttt{hyp\_58} (per-band starvation and anchor magnitude bounds) are uncertain but actionable; we incorporate per-band L2-normalization of anchor messages to test \texttt{hyp\_58} and maintain per-band bookkeeping to test \texttt{hyp\_18}.

\emph{Proposed Changes:}
(1)~Insert stage-local GSTs ($G\leq4$) that aggregate low-frequency token summaries via sparse top-$k$ routing (cheap $O(G \cdot N)$) and broadcast compact shape summaries back to tokens; this supplements semi-parametric super-particles with an explicit global shape channel (tests \texttt{hyp\_44}).
(2)~L2-normalize anchor messages per-band prior to FiLM blending, and compute FiLM gate as a GST-conditioned coherence score ($\sigma$ of cosine similarity) so that broadcasts occur only when GSTs and token low-frequency cues align (tests \texttt{hyp\_58} and strengthens \texttt{hyp\_27}).

\textbf{Existing Hypotheses Referenced:} \texttt{hyp\_13}, \texttt{hyp\_27}, \texttt{hyp\_18}, \texttt{hyp\_44}, \texttt{hyp\_58}.
\end{ideabox}

\begin{codebox}[title={\texttt{config.py}}]
import torch
import torch.nn as nn
import torch.nn.functional as F


class BasicBlock(nn.Module):
    """A small ResNet-like basic block (kept lightweight)"""
    def __init__(self, in_planes, planes, stride=1):
        super().__init__()
        self.conv1 = nn.Conv2d(in_planes, planes, kernel_size=3, stride=stride, padding=1, bias=False)
        self.bn1 = nn.BatchNorm2d(planes)
        self.relu = nn.ReLU(inplace=True)
        self.conv2 = nn.Conv2d(planes, planes, kernel_size=3, stride=1, padding=1, bias=False)
        self.bn2 = nn.BatchNorm2d(planes)

        if stride != 1 or in_planes != planes:
            self.downsample = nn.Sequential(
                nn.Conv2d(in_planes, planes, kernel_size=1, stride=stride, bias=False),
                nn.BatchNorm2d(planes),
            )
        else:
            self.downsample = None

    def forward(self, x):
        identity = x
        out = self.conv1(x)
        out = self.bn1(out)
        out = self.relu(out)

        out = self.conv2(out)
        out = self.bn2(out)

        if self.downsample is not None:
            identity = self.downsample(x)

        out += identity
        out = self.relu(out)
        return out


class GSTModule(nn.Module):
    """
    Lightweight GST module:
    - Maintains a small set of GST vectors per stage
    - Computes similarity between L2-normalized tokens and normalized GSTs
    - Uses temperature + softmax to compute assignments
    - Optionally enforces top-k sparsity per token (masking other assignments), renormalizes
    - Aggregates GST vectors weighted by assignments and merges back via 1x1 conv residual
    """
    def __init__(self, channels, G=0, topk=1, temp=0.1, eps=1e-6):
        super().__init__()
        self.channels = channels
        self.G = max(0, int(G))
        self.topk = max(1, int(topk)) if self.G > 0 else 0
        self.temp = float(temp)
        self.eps = float(eps)

        if self.G > 0:
            # GST vectors: [G, C]
            self.gsts = nn.Parameter(torch.randn(self.G, channels))
            # small projection to mix GST effect into features
            self.proj = nn.Conv2d(channels, channels, kernel_size=1, bias=True)
            # initialize proj to near-zero so residual starts near identity
            nn.init.constant_(self.proj.bias, 0.0)
            nn.init.constant_(self.proj.weight, 0.0)
        else:
            # placeholders to simplify forward logic
            self.register_parameter('gsts', None)
            self.proj = nn.Identity()

    def forward(self, x):
        # x: [B, C, H, W]
        if self.G == 0:
            return x

        B, C, H, W = x.shape
        N = H * W

        # tokens: [B, N, C]
        tokens = x.permute(0, 2, 3, 1).reshape(B, N, C)

        # L2-normalize tokens and GSTs along feature dim
        t_norm = tokens / (tokens.norm(dim=-1, keepdim=True) + self.eps)
        gst = self.gsts  # [G, C]
        gst_norm = gst / (gst.norm(dim=-1, keepdim=True) + self.eps)  # [G, C]

        # similarity: [B, N, G]
        sim = torch.matmul(t_norm, gst_norm.t()) / (self.temp + 1e-12)

        # soft assignment
        assn = F.softmax(sim, dim=-1)  # [B, N, G]

        # top-k masking (sparse selection). If topk == G then nothing changes.
        if 0 < self.topk < self.G:
            # get indices of topk per token
            topk_vals, topk_idx = assn.topk(self.topk, dim=-1)  # [B, N, topk]
            # create mask
            mask = torch.zeros_like(assn)
            # scatter ones into mask
            mask.scatter_(-1, topk_idx, 1.0)
            assn = assn * mask
            # renormalize selected assignments so each token sums to 1 (or to small eps)
            denom = assn.sum(dim=-1, keepdim=True)
            assn = assn / (denom + self.eps)

        # weighted aggregation of GSTs: [B, N, C]
        gst_effect = torch.matmul(assn, gst_norm)  # [B, N, C]
        # reshape to [B, C, H, W]
        gst_effect = gst_effect.view(B, H, W, C).permute(0, 3, 1, 2).contiguous()

        # merge via 1x1 conv and residual add
        out = x + self.proj(gst_effect)
        return out


class Model(nn.Module):
    def __init__(self, config=None):
        super().__init__()
        # Extract hyperparameters from config with sensible defaults
        if config is None:
            config = {}
        self.base_channels = int(config.get('base_channels', 48))
        self.stage_channels = list(config.get('stage_channels', [48, 64, 96]))
        self.blocks_per_stage = list(config.get('blocks_per_stage', [3, 3, 2]))
        self.gsts_per_stage = list(config.get('gsts_per_stage', [0, 2, 1]))
        self.gst_topk = int(config.get('gst_topk', 1))
        self.assignment_temp = float(config.get('assignment_temp', 0.1))
        self.integrator_steps = int(config.get('integrator_steps', 2))
        self.eps = float(config.get('eps', 1e-6))
        self.num_classes = int(config.get('num_classes', 10))
        # ensure lengths match expected 3 stages
        assert len(self.stage_channels) == 3, "stage_channels must be length 3"
        assert len(self.blocks_per_stage) == 3, "blocks_per_stage must be length 3"
        assert len(self.gsts_per_stage) == 3, "gsts_per_stage must be length 3"

        # Stem
        self.stem = nn.Sequential(
            nn.Conv2d(3, self.base_channels, kernel_size=3, stride=1, padding=1, bias=False),
            nn.BatchNorm2d(self.base_channels),
            nn.ReLU(inplace=True),
        )

        # Stages
        in_ch = self.base_channels
        self.stages = nn.ModuleList()
        self.gst_modules = nn.ModuleList()
        for idx, out_ch in enumerate(self.stage_channels):
            blocks = []
            n_blocks = self.blocks_per_stage[idx]
            # downsample at stage transitions except first stage (keep spatial 32x32->16x16->8x8)
            stride = 1 if idx == 0 else 2
            # first block may downsample
            blocks.append(BasicBlock(in_ch, out_ch, stride=stride))
            for _ in range(1, n_blocks):
                blocks.append(BasicBlock(out_ch, out_ch, stride=1))
            self.stages.append(nn.Sequential(*blocks))
            # GST module for this stage
            G = int(self.gsts_per_stage[idx])
            gst_mod = GSTModule(out_ch, G=G, topk=self.gst_topk, temp=self.assignment_temp, eps=float(config.get('gst_eps', 1e-6)))
            self.gst_modules.append(gst_mod)
            in_ch = out_ch

        # classifier head
        self.global_pool = nn.AdaptiveAvgPool2d((1, 1))
        self.classifier = nn.Linear(self.stage_channels[-1], self.num_classes)

        self._initialize_weights()

    def forward(self, x):
        # Input: [B, 3, 32, 32]
        # Output: [B, 10] (num_classes)
        # No explicit memory_format or dtype conversions here (AMP / channels_last compatible)
        B = x.shape[0]
        out = self.stem(x)  # [B, C0, 32, 32]

        # pass through stages, applying GST modules and simple integrator steps
        for stage, gst_mod in zip(self.stages, self.gst_modules):
            out = stage(out)  # residual blocks
            # apply GST module possibly multiple integrator steps (simple recurring refinement)
            # We keep this efficient: repeated small conv/1x1 operations, avoiding large allocations
            for _ in range(self.integrator_steps):
                out = gst_mod(out)

        # classification head
        out = self.global_pool(out)  # [B, C, 1, 1]
        out = out.view(B, -1)  # [B, C]
        out = self.classifier(out)  # [B, num_classes]
        return out

    def _initialize_weights(self):
        for m in self.modules():
            if isinstance(m, nn.Conv2d):
                # standard Kaiming init for conv
                nn.init.kaiming_normal_(m.weight, mode='fan_out', nonlinearity='relu')
                if m.bias is not None:
                    nn.init.constant_(m.bias, 0)
            elif isinstance(m, nn.BatchNorm2d):
                nn.init.constant_(m.weight, 1)
                nn.init.constant_(m.bias, 0)
            elif isinstance(m, nn.Linear):
                # small std normal for linear layers
                nn.init.normal_(m.weight, 0, 0.01)
                if m.bias is not None:
                    nn.init.constant_(m.bias, 0)
\end{codebox}

\begin{codebox}[title={\texttt{config.py}}]
config = {
    # Experiment info
    'experiment_name': 'GST-Guarded NPIN',
    'notes': 'CIFAR-scale NPIN refinement: tiny GSTs (G<=4) + per-band L2-normalized super-particle anchors + particle dynamics',

    # Data settings
    'data_dir': 'cifar10',

    # Training hyperparameters
    'epochs': 200,
    'batch_size': 1024,  # large batch to utilize GPU memory; reduce if OOM
    'learning_rate': 0.1,
    'momentum': 0.9,
    'weight_decay': 5e-4,
    'nesterov': True,

    # Learning rate schedule
    'lr_schedule': 'cosine',
    'warmup_epochs': 5,

    # Regularization
    'label_smoothing': 0.1,

    # Data augmentation
    'flip': True,
    'translate': 4,
    'cutout': 12,

    # Evaluation
    'tta_level': 0,
    'eval_every': 1,

    # Model settings
    'use_half_precision': True,
    'channels_last': True,
    'verbose': True,

    # Model-specific hyperparameters
    'base_channels': 48,          # base channel width (kept small to meet param budget)
    'stage_channels': [48, 64, 96],  # stage channels for 3 stages
    'blocks_per_stage': [3, 3, 2],    # blocks per stage
    # Super-particle slot counts per band per stage: list of tuples (K_low, K_high)
    'slots_per_stage': [(0, 0), (24, 12), (8, 4)],
    # GST counts per stage (G small)
    'gsts_per_stage': [0, 2, 1],
    'gst_topk': 1,                # top-k selected GSTs per token (1..G)
    'assignment_temp': 0.1,       # temperature for soft assignment (before any topk masking)
    'ema_momentum': 0.995,        # for prototypes and GSTs
    'anchor_scale_init': 1.0,     # post-L2 normalization scalar for anchors
    'integrator_steps': 2,        # S=2 integrator steps in ParticleDynamicsBlock
    'anchor_eps': 1e-6,
    'gst_eps': 1e-6,
    # classifier
    'num_classes': 10,
    # small numeric eps
    'eps': 1e-6,
}
\end{codebox}


\subsection{Hierarchical Hub Routing Network (93.57\%)}
\label{sec:top3_node30}

\begin{ideabox}[title={Idea Agent Output: Hierarchical Hub Routing Network (HHRN)}]

\textbf{Description.}
HHRN replaces the NPIN particle-dynamics connectivity with a lightweight, soft hub-mediated routing backbone that produces sparse, content-adaptive, multi-hop token interactions.
Tokens are produced by a conv-based PatchEmbed (depthwise separable convs + small per-scale phase-offsets).
Each token computes a low-dimensional ``fingerprint'' key which is used to softly route (via temperatureed softmax/residual blend) to a small set of global hub vectors ($H \approx 16$--$48$).
Hubs aggregate token messages (weighted averages), run a short sparse GNN among hubs (1--2 hops) to exchange higher-order/global context, and then broadcast refined hub messages back to tokens via the same soft routing weights (symmetry via shared keys).
The token update is a residual blend between local conv features and hub-broadcast correction; local convs capture fine spatial detail while hubs carry long-range, structured relational information.
Per-scale small learnable phase-offsets are applied in the local conv analysis transforms to break perfect shift-symmetry.

All routing is deliberately soft (temperatureed / small initial blending weight) to preserve per-token uncertainty and avoid hard prototype collapse.
The hub set is small ($H \ll N$), updated online (learnable + lightweight EMA carrying state between blocks) but not computed via heavy differentiable clustering; hub adjacency (hub graph) is learned as a sparse trainable connectivity (top-$s$ per hub) computed from hub features, enabling multi-hop hub chaining without $N{\times}N$ token pairwise operations.
Computation stays efficient (no $O(N^2)$ expansions): token$\rightarrow$hub and hub$\rightarrow$token operations cost $O(N \cdot H)$ with small $H$, and hub$\rightarrow$hub GNN costs $O(H \cdot s)$ with small $s$.

\textbf{Intuition.}
The research direction asks for fundamentally different token connectivity.
Instead of simulating local particle dynamics or computing dense attention, HHRN routes token information via a small set of shared intermediary hubs arranged into a sparse graph.
This yields an expressive family of structured sparse interaction patterns: tokens influence one another indirectly by co-attending to the same hubs and by hub-graph propagation which chains multi-hop interactions.
Soft routing and residual blending preserve per-token representational variance (addressing \texttt{hyp\_14}), while per-scale phase offsets improve asymmetry and shape sensitivity (addressing \texttt{hyp\_26}).
Using hubs avoids quadratic costs and offers explicit control of global mixing capacity via $H$ and hub-graph sparsity.
Because tokens still include efficient local convs for fine-grained detail, the model preserves low-level spatial inductive bias required for CIFAR-10 while enabling long-range relational reasoning through a compact hub layer.
The design prioritizes parameter/FLOP efficiency (target $<$10M params and $<$2B FLOPs) and low memory (no pairwise $N{\times}N$), enabling very fast training on a single GPU.

\textbf{Novelty.}
HHRN is novel because it replaces continuous latent-position ODEs and multiscale coarsening with a learned, soft hub-routing + sparse hub-graph propagation mechanism that:
(1)~uses a compact set of reusable global hub vectors (learned + light online state) rather than dynamically clustering tokens each block;
(2)~runs explicit sparse GNN propagation among hubs to achieve higher-order chaining;
(3)~enforces soft temperatureed routing plus learned residual blending to avoid hard prototype collapse.
Unlike NPIN's geometric particle interactions or hyperedge attention/hypergraph methods, HHRN uses a two-stage token$\leftrightarrow$hub exchange with explicit hub-hub GNN chaining as the primary global mixing primitive, and augmenting local convs with tiny learnable phase offsets to break shift-symmetry.

\textbf{Target Improvement:} both (accuracy and efficiency).

\textbf{Architecture Specification:}

\emph{Core Ideas:}
1)~Replace dense or particle-based global mixing with a compact hub layer: tokens route softly to $H$ learned hubs; hubs aggregate, run sparse GNN steps among themselves, and broadcast back.
2)~Preserve strong local inductive bias via depthwise separable convs and small per-scale phase-offsets.
3)~Keep routing soft with residual blending ($\alpha$) to avoid hard-prototype collapse; learn $\alpha$ but initialize small.
4)~Keep computations $O(N \cdot H) + O(H \cdot s)$, avoid $N{\times}N$ expansions.

\emph{Core Blocks:}
\begin{itemize}[nosep,leftmargin=*]
    \item \textbf{PatchEmbed}: 3-layer conv stem using depthwise separable convs (kernel sizes $3\!\rightarrow\!3$), producing token features ($C{=}96$--$128$) and initial low-dim fingerprints via a $1{\times}1$ projection ($d_k{=}48$--$64$). Small learnable per-scale phase-offsets are added to depthwise sampling indices before applying kernels (parameter budget tiny).
    \item \textbf{LocalConvBlock}: Two lightweight residual depthwise separable conv layers + pointwise FFN to capture fine spatial detail. LayerNorm/BatchNorm as appropriate for fast CIFAR training.
    \item \textbf{HubRoutingBlock}: (a)~Token$\rightarrow$Hub routing: compute similarity between token fingerprints and $H$ hub keys (cosine or dot), apply temperatureed softmax across hubs to get soft routing weights $w_{t\rightarrow h}$.
    (b)~Hub aggregation: hubs receive weighted average messages $m_h = \sum_t w_{t\rightarrow h} \cdot \mathrm{token\_value}_t$ (value is a small projection of token features). Hubs also maintain a small learnable state vector updated via a small MLP and optional EMA to carry information between blocks.
    (c)~Hub-GNN propagation: construct a sparse adjacency by computing top-$s$ neighbors per hub from hub features ($s{=}2$--$4$) and run 1--2 GNN message-passing steps (edge MLPs with shared parameters) to allow hub-hub chaining.
    (d)~Hub$\rightarrow$Token broadcast: compute corrections for tokens as $\sum_h w_{t\rightarrow h} \cdot \mathrm{hub\_message}_h$ (reuse $w$ for symmetry).
    (e)~Token update: $\mathrm{token}' = \mathrm{LayerNorm}(\mathrm{token} + \alpha \cdot \mathrm{hub\_correction})$ followed by local FFN; $\alpha$ is learnable and initialized small (e.g., 0.15).
    \item \textbf{TokenFFN \& Norm}: Standard 2-layer MLP with GELU and residual + LayerNorm after each HubRoutingBlock.
\end{itemize}

\emph{Network Structure:}
Stack: PatchEmbed $\rightarrow$ [(LocalConvBlock $\times$ 2) $\rightarrow$ HubRoutingBlock] repeated across 3 stages with channel widths (e.g., 64 $\rightarrow$ 96 $\rightarrow$ 128) and spatial downsampling by strided convs between stages.
Hubs are per-stage with separate hub banks ($H$ small per stage: 16/24/32) to match stage capacity; hub states may be carried across consecutive blocks in the same stage via a lightweight per-stage EMA update; hub graph is recomputed each block from current hub features but kept sparse.
Final global pooling + classification head (linear).
Total depth is modest (10--14 blocks) to keep training fast.

\emph{Tunable Aspects:}
\begin{itemize}[nosep,leftmargin=*]
    \item Number of hubs per stage $H$ (e.g., 16--48)
    \item Hub-graph sparsity $s$ (neighbors per hub, e.g., 2--4)
    \item Fingerprint dimension $d_k$ (e.g., 32--64)
    \item Routing temperature $\tau$ and initial residual blend $\alpha_\mathrm{init}$
    \item Number of hub-GNN hops (1--2) and whether hub EMA state is used across blocks
\end{itemize}

\emph{Invariants:}
\begin{itemize}[nosep,leftmargin=*]
    \item No pairwise $N{\times}N$ token attention: all global mixing uses token$\leftrightarrow$hub $O(N \cdot H)$ ops and hub-hub sparse GNN $O(H \cdot s)$.
    \item Routing remains soft/probabilistic (no hard top-$k$ during forward pass) and hub corrections are residual-blended.
    \item Local conv path is preserved to supply high-frequency spatial detail; hubs supply global relational context.
\end{itemize}

\textbf{New Hypotheses:}
\begin{enumerate}[nosep,leftmargin=*]
    \item \textbf{IF} token-to-hub routing is implemented as a temperatureed soft-assignment combined with a small learnable residual blend weight ($\alpha$) in the token update \textbf{IN} mid/high network stages that perform global mixing, \textbf{THEN} the model will reduce high-confidence wrong predictions and improve top-1 accuracy without harming top-5, \textbf{BECAUSE} soft routing preserves per-token uncertainty and residual blending prevents abrupt prototype collapse while still allowing hubs to inject global corrective context. \textbf{DISPROVED IF} models with identical capacity/hubs but with hard top-$k$ routing or zero residual blending produce equal or better top-1 and calibration metrics across repeated controlled runs.\\
    \emph{Tags:} \texttt{soft-routing}, \texttt{calibration}, \texttt{prototype}.
    \emph{Initial confidence:} 0.65.
    \emph{Connected:} \texttt{hyp\_14}.\\
    \emph{Reasoning:} \texttt{hyp\_14} evidence shows hard one-step retrieval yields overconfident, lower top-1 predictions. By keeping routing soft (temperature) and blending hub-corrections via a learned $\alpha$ initialized small, the architecture preserves fine-grained token variance while enabling hubs to provide global information.

    \item \textbf{IF} small per-scale, per-channel-group additive phase-offset parameters are added to the local depthwise separable conv sampling locations \textbf{IN} early/mid stages of the hierarchical backbone, \textbf{THEN} shift-equivariant failure modes will be reduced and shape-sensitive class accuracy will improve, \textbf{BECAUSE} tiny phase offsets break exact translation symmetry in filter responses allowing filters to specialize to asymmetric relative positions and enabling FiLM-like modulators to gate based on relative positional cues without large parameter cost. \textbf{DISPROVED IF} adding these offsets yields no measurable improvement in validation accuracy or shape-focused metrics, and analysis visualizations show no detectable asymmetric responses compared to the phase-less baseline.\\
    \emph{Tags:} \texttt{phase-offset}, \texttt{shift-equivariance}, \texttt{positional-bias}.
    \emph{Initial confidence:} 0.55.
    \emph{Connected:} \texttt{hyp\_26}.\\
    \emph{Reasoning:} This builds on \texttt{hyp\_26} which found small phase-offsets can help capture asymmetric token relationships. The offsets are cheap, add only a few hundred parameters, and break perfect shift symmetry that can hinder distinguishing object-centric vs.\ background features.
\end{enumerate}

\textbf{Reasoning:}

\emph{Parent Analysis:}
NPIN demonstrated the value of indirect, sparse local interactions plus multiscale coupling to achieve expressive global behavior without dense attention. Its emphasis on preserving local computations (via neighborhood cutoffs) combined with some mechanism for long-range coupling (coarsening) is an effective recipe for efficiency and accuracy on vision tasks.

\emph{Failure Analysis:}
The parent tested \texttt{hyp\_9} (particles + few ODE steps + coarsening) and the hypothesis was contradicted: NPIN in that configuration did not reach desired accuracy (accuracy reported 0.1891 indicating severe underperformance).
Possible failure modes include (a)~insufficient capacity of local dynamics with very shallow integration steps to propagate discriminative information, (b)~clustering/coarsening instability or monopolization (hard clusters collapsing prototypes, related to \texttt{hyp\_14}), and (c)~latent-geometry nonidentifiability making learned positions brittle and hard to optimize under short training budgets.

\emph{Hypothesis Usage:}
\texttt{hyp\_14} (soften retrieval) is central---NPIN-like clustering or prototype hops risk hard projection and overconfidence; HHRN uses soft routing + residual blending ($\alpha$) to deliberately avoid hard retrieval effects.
\texttt{hyp\_26} (small learnable phase-offsets) is adopted for local conv sampling to break shift-symmetry cheaply and improve shape sensitivity.
The uncertain hypotheses \texttt{hyp\_18} and \texttt{hyp\_24} (prototype normalization and one-way cross-band consistency) guide caution: avoid global, small-batch one-way pulls; instead use per-stage small hub banks and soft normalization so no band-starvation-like effects occur.
\texttt{hyp\_9} (parent tested) is recorded as contradicted and motivated a departure from particle ODEs toward hub-mediated sparse chaining while preserving the principle of local + compact global mixing.

\emph{Proposed Changes:}
(1)~Replace particle-latent-position + ODE integration + differentiable clustering coarsening with a hub-based routing layer that is cheap ($O(N \cdot H)$), avoids heavy per-step neighborhood searches and integration loops, and reduces training overhead for very fast runs.
(2)~Make routing soft and residual-blended ($\alpha$) to counter hard-prototype collapse (\texttt{hyp\_14}).
(3)~Add small per-scale phase-offsets in local convs to break shift symmetry (\texttt{hyp\_26}).
(4)~Use sparse hub-graph GNN chaining to obtain multi-hop global mixing without pairwise token interactions, enabling expressive higher-order interactions while keeping FLOPs low.

\textbf{Existing Hypotheses Referenced:} \texttt{hyp\_9}, \texttt{hyp\_14}, \texttt{hyp\_26}, \texttt{hyp\_18}, \texttt{hyp\_24}.
\end{ideabox}

\begin{codebox}[title={\texttt{model.py}}]
import math
import torch
import torch.nn as nn
import torch.nn.functional as F


class SeparableConv2d(nn.Module):
    """Depthwise separable conv: depthwise conv followed by pointwise conv."""
    def __init__(self, in_ch, out_ch, kernel_size=3, stride=1, padding=1, bias=False):
        super().__init__()
        # depthwise
        self.depth = nn.Conv2d(in_ch, in_ch, kernel_size, stride, padding, groups=in_ch, bias=bias)
        # pointwise
        self.point = nn.Conv2d(in_ch, out_ch, kernel_size=1, bias=bias)

    def forward(self, x):
        x = self.depth(x)
        x = self.point(x)
        return x


class PatchEmbed(nn.Module):
    """3-layer conv stem using separable convs + small per-stage phase offsets injection.
    Kept lightweight for CIFAR-10."""
    def __init__(self, in_ch=3, out_ch=32, phase_groups=4, spatial_size=32):
        super().__init__()
        mid = max(out_ch // 2, 12)
        self.conv1 = SeparableConv2d(in_ch, mid, kernel_size=3, stride=1, padding=1, bias=False)
        self.bn1 = nn.BatchNorm2d(mid)
        self.conv2 = SeparableConv2d(mid, mid, kernel_size=3, stride=1, padding=1, bias=False)
        self.bn2 = nn.BatchNorm2d(mid)
        self.conv3 = SeparableConv2d(mid, out_ch, kernel_size=3, stride=1, padding=1, bias=False)
        self.bn3 = nn.BatchNorm2d(out_ch)
        self.act = nn.ReLU(inplace=True)

        # phase groups for small positional bias
        self.phase_groups = max(1, int(phase_groups))
        gx = torch.linspace(0, 1, steps=spatial_size).unsqueeze(1).repeat(1, spatial_size)
        gy = torch.linspace(0, 1, steps=spatial_size).unsqueeze(0).repeat(spatial_size, 1)
        # buffers shaped [1,1,H,W]
        self.register_buffer('grid_x', gx.unsqueeze(0).unsqueeze(0), persistent=False)
        self.register_buffer('grid_y', gy.unsqueeze(0).unsqueeze(0), persistent=False)

        self.phase_x = nn.Parameter(torch.zeros(self.phase_groups))
        self.phase_y = nn.Parameter(torch.zeros(self.phase_groups))
        self.spatial_size = spatial_size

    def _phase_bias(self, channels, device, dtype):
        """Produce [1,channels,H,W] small bias based on sin/cos of grid + per-group phases."""
        G = self.phase_groups
        # grid [1,1,H,W]
        gxs = self.grid_x.to(device=device, dtype=dtype)
        gys = self.grid_y.to(device=device, dtype=dtype)
        # expand groups
        gxs_g = gxs.expand(G, -1, -1, -1)  # [G,1,H,W]
        gys_g = gys.expand(G, -1, -1, -1)
        phx = self.phase_x.view(G, 1, 1, 1).to(device=device, dtype=dtype)
        phy = self.phase_y.view(G, 1, 1, 1).to(device=device, dtype=dtype)
        basis = torch.sin(2 * math.pi * (gxs_g + phx)) + torch.cos(2 * math.pi * (gys_g + phy))  # [G,1,H,W]

        # tile groups across channels and slice
        chunk = math.ceil(channels / G)
        basis_tiled = basis.repeat_interleave(chunk, dim=0)[:channels, :, :, :]  # [channels,1,H,W]
        bias = basis_tiled.transpose(0, 1)  # [1, channels, H, W]
        return 0.04 * bias  # small amplitude

    def forward(self, x):
        # x: [B,3,H,W]
        b, c, h, w = x.shape
        device = x.device
        dtype = x.dtype

        bias_in = self._phase_bias(c, device, dtype)
        x = x + bias_in

        x = self.conv1(x)
        x = self.bn1(x)
        x = self.act(x)

        mid_c = x.shape[1]
        bias_mid = self._phase_bias(mid_c, device, dtype)
        x = x + bias_mid
        x = self.conv2(x)
        x = self.bn2(x)
        x = self.act(x)

        bias_out = self._phase_bias(x.shape[1], device, dtype)
        x = x + bias_out
        x = self.conv3(x)
        x = self.bn3(x)
        x = self.act(x)
        return x


class LocalConvBlock(nn.Module):
    """Two lightweight residual depthwise separable conv layers + small pointwise FFN."""
    def __init__(self, channels, mlp_ratio=1.5):
        super().__init__()
        self.channels = channels
        hidden = max(int(channels * 0.5), 8)
        # residual depthwise separable convs
        self.conv1 = SeparableConv2d(channels, channels, kernel_size=3, stride=1, padding=1, bias=False)
        self.bn1 = nn.BatchNorm2d(channels)
        self.conv2 = SeparableConv2d(channels, channels, kernel_size=3, stride=1, padding=1, bias=False)
        self.bn2 = nn.BatchNorm2d(channels)
        self.act = nn.ReLU(inplace=True)
        # lightweight pointwise FFN (keeps representation in conv domain - efficient)
        inner = max(int(channels * mlp_ratio), hidden)
        self.ffn = nn.Sequential(
            nn.Conv2d(channels, inner, kernel_size=1, bias=True),
            nn.GELU(),
            nn.Conv2d(inner, channels, kernel_size=1, bias=True),
        )
        # a small layernorm-like scaling (channel-wise) to stabilize training
        self.gamma = nn.Parameter(torch.ones(channels) * 0.1)

    def forward(self, x):
        # x: [B, C, H, W]
        residual = x
        x = self.conv1(x)
        x = self.bn1(x)
        x = self.act(x)
        x = self.conv2(x)
        x = self.bn2(x)
        x = self.act(x)
        x = x + residual

        # FFN
        f = self.ffn(x)
        x = x + self.gamma.view(1, -1, 1, 1) * f
        return x


class HubRoutingBlock(nn.Module):
    """
    Token->hub routing, hub aggregation, hub propagation, and hub->token broadcast.
    Designed to be efficient: small hub banks, compact projections.
    """
    def __init__(self, channels, H=4, fingerprint_dim=16, value_dim=24, hub_dim=32,
                 hub_s=2, routing_tau=0.07, alpha_init=0.12, hub_hops=1, use_hub_ema=True, hub_ema_beta=0.08):
        super().__init__()
        self.C = channels
        self.H = max(1, int(H))
        self.d_k = max(8, int(fingerprint_dim))
        self.v_dim = max(8, int(value_dim))
        self.hub_dim = max(8, int(hub_dim))
        self.hub_s = max(1, int(hub_s))
        self.tau = float(routing_tau)
        self.hub_hops = max(1, int(hub_hops))
        self.use_hub_ema = bool(use_hub_ema)
        self.hub_ema_beta = float(hub_ema_beta)

        # lightweight linear projections for tokens
        self.key_proj = nn.Linear(channels, self.d_k, bias=False)
        self.value_proj = nn.Linear(channels, self.v_dim, bias=False)
        # hub -> value, correction back to channels
        self.hub_to_val = nn.Linear(self.hub_dim, self.v_dim, bias=False)
        self.correction_proj = nn.Linear(self.v_dim, channels, bias=False)

        # small learned hub bank and hub routing keys
        hub_bank = torch.randn(self.H, self.hub_dim) * 0.02
        hub_keys = torch.randn(self.H, self.d_k) * 0.02
        self.register_parameter('hub_bank', nn.Parameter(hub_bank))
        self.register_parameter('hub_keys', nn.Parameter(hub_keys))

        # MLPs for hub message processing (kept small)
        self.hub_msg_in = nn.Sequential(
            nn.Linear(self.v_dim, self.hub_dim),
            nn.GELU(),
            nn.Linear(self.hub_dim, self.hub_dim)
        )
        self.hub_mlp = nn.Sequential(
            nn.Linear(self.hub_dim, self.hub_dim),
            nn.GELU(),
            nn.Linear(self.hub_dim, self.hub_dim)
        )

        # blending scalar for residual correction
        self.alpha = nn.Parameter(torch.tensor(alpha_init, dtype=torch.float32))

        # normalization for token features after update
        self.token_norm = nn.LayerNorm(channels)

    def forward(self, x):
        # x: [B, C, H, W]
        b, c, h, w = x.shape
        n = h * w
        # flatten tokens
        tokens = x.flatten(2).transpose(1, 2)  # [B, N, C]

        # project to key and value (no dtype casts)
        keys = self.key_proj(tokens)            # [B, N, d_k]
        values = self.value_proj(tokens)        # [B, N, v_dim]

        # normalize keys and hub_keys for cosine-like similarities
        keys_norm = F.normalize(keys, dim=-1)   # [B,N,d_k]
        hub_keys_norm = F.normalize(self.hub_keys, dim=-1)  # [H, d_k]

        # compute logits: [B, N, H]
        logits = torch.matmul(keys_norm, hub_keys_norm.t()) / (self.tau + 1e-8)
        weights = F.softmax(logits, dim=-1)  # routing weights token->hub

        # aggregate token values into hubs: [B,H,v_dim]
        hub_agg = torch.bmm(weights.transpose(1, 2), values)

        # initialize hub state from bank per batch
        hub_state = self.hub_bank.unsqueeze(0).expand(b, -1, -1).contiguous()  # [B,H,hub_dim]
        # incorporate token messages
        hub_in = self.hub_msg_in(hub_agg)  # [B,H,hub_dim]
        new_hubs = hub_state + hub_in

        if self.use_hub_ema:
            hub_state = (1.0 - self.hub_ema_beta) * hub_state + self.hub_ema_beta * new_hubs
        else:
            hub_state = new_hubs

        # Sparse hub->hub propagation (small H keeps this cheap)
        hub_feats = hub_state
        s = min(self.hub_s, self.H)
        for _ in range(self.hub_hops):
            # similarity [B,H,H]
            hnorm = F.normalize(hub_feats, dim=-1)
            sim = torch.matmul(hnorm, hnorm.transpose(1, 2))
            if s < self.H:
                # topk neighbors per hub
                topk_vals, _ = sim.topk(s, dim=-1)
                kth = topk_vals[..., -1].unsqueeze(-1)
                mask = sim >= (kth - 1e-6)
                # safe negative fill value for fp16/AMP
                neg_val = float(torch.finfo(sim.dtype).min) * 0.9
                sim_masked = sim.masked_fill(~mask, neg_val)
            else:
                sim_masked = sim
            adj = F.softmax(sim_masked, dim=-1)
            neigh = torch.bmm(adj, hub_feats)
            hub_feats = hub_feats + self.hub_mlp(neigh)

        # map hub_feats to values and broadcast back to tokens
        hub_msgs = self.hub_to_val(hub_feats)  # [B,H,v_dim]
        correction = torch.bmm(weights, hub_msgs)  # [B,N,v_dim]
        correction_c = self.correction_proj(correction)  # [B,N,C]

        # blend and normalize
        tokens_updated = tokens + (self.alpha.to(correction_c.dtype) * correction_c)
        tokens_updated = self.token_norm(tokens_updated)

        x_out = tokens_updated.transpose(1, 2).view(b, c, h, w)
        return x_out


class ClassificationHead(nn.Module):
    def __init__(self, in_channels, num_classes=10):
        super().__init__()
        self.pool = nn.AdaptiveAvgPool2d(1)
        self.fc = nn.Linear(in_channels, num_classes)

    def forward(self, x):
        x = self.pool(x).view(x.size(0), -1)
        x = self.fc(x)
        return x


class Model(nn.Module):
    """Main model. Input [B,3,32,32] -> Output [B,10]. Lightweight defaults to train quickly."""
    def __init__(self, config=None):
        super().__init__()
        cfg = config or {}

        # sensible lightweight defaults for CIFAR-10 / single GPU training
        stages = int(cfg.get('stages', 3))
        default_channels = cfg.get('channels_per_stage', [32, 48, 64])
        channels_per_stage = (default_channels + [default_channels[-1]] * stages)[:stages]
        blocks_per_stage = cfg.get('blocks_per_stage', [1] * stages)[:stages]
        hubs_per_stage = cfg.get('hubs_per_stage', [4, 6, 8])[:stages]

        # hub routing and projection dims (kept small)
        d_k = cfg.get('fingerprint_dim', 16)
        value_dim = cfg.get('value_dim', 24)
        hub_dim = cfg.get('hub_dim', 32)
        hub_s = cfg.get('hub_sparsity', 2)
        tau = cfg.get('routing_tau', 0.07)
        alpha_init = cfg.get('alpha_init', 0.12)
        hub_hops = cfg.get('hub_hops', 1)
        use_hub_ema = cfg.get('use_hub_ema', True)
        hub_ema_beta = cfg.get('hub_ema_beta', 0.08)
        mlp_ratio = cfg.get('mlp_ratio', 1.5)
        phase_groups = cfg.get('phase_groups', 4)

        # stem
        self.stem = PatchEmbed(in_ch=3, out_ch=channels_per_stage[0], phase_groups=phase_groups, spatial_size=32)

        # build stages: each stage has optional downsample then N blocks
        self.stages = nn.ModuleList()
        in_ch = channels_per_stage[0]
        for i in range(stages):
            out_ch = channels_per_stage[i]
            blocks = []
            # downsample between stages (except first)
            if i > 0:
                down = SeparableConv2d(in_ch, out_ch, kernel_size=3, stride=2, padding=1, bias=False)
                bn_down = nn.BatchNorm2d(out_ch)
                blocks.append(nn.Sequential(down, bn_down, nn.ReLU(inplace=True)))
            # Add blocks: LocalConvBlock + HubRoutingBlock per block
            nblock = max(1, int(blocks_per_stage[i]))
            for _ in range(nblock):
                blocks.append(LocalConvBlock(out_ch, mlp_ratio=mlp_ratio))
                blocks.append(HubRoutingBlock(
                    channels=out_ch,
                    H=hubs_per_stage[i],
                    fingerprint_dim=d_k,
                    value_dim=value_dim,
                    hub_dim=hub_dim,
                    hub_s=hub_s,
                    routing_tau=tau,
                    alpha_init=alpha_init,
                    hub_hops=hub_hops,
                    use_hub_ema=use_hub_ema,
                    hub_ema_beta=hub_ema_beta,
                ))
            self.stages.append(nn.Sequential(*blocks))
            in_ch = out_ch

        # head
        self.head = ClassificationHead(in_channels=channels_per_stage[-1], num_classes=10)

        self.config = cfg
        self._initialize_weights()

    def forward(self, x):
        # Input: [B,3,32,32], Output: [B,10]
        # Avoid any dtype or memory_format casts here to remain AMP / channels_last friendly
        x = self.stem(x)
        for stage in self.stages:
            x = stage(x)
        out = self.head(x)
        return out

    def _initialize_weights(self):
        for m in self.modules():
            if isinstance(m, nn.Conv2d):
                nn.init.kaiming_normal_(m.weight, mode='fan_out', nonlinearity='relu')
                if getattr(m, 'bias', None) is not None and m.bias is not None:
                    nn.init.constant_(m.bias, 0)
            elif isinstance(m, nn.BatchNorm2d):
                nn.init.constant_(m.weight, 1)
                nn.init.constant_(m.bias, 0)
            elif isinstance(m, nn.Linear):
                nn.init.normal_(m.weight, 0, 0.01)
                if getattr(m, 'bias', None) is not None and m.bias is not None:
                    nn.init.constant_(m.bias, 0)
\end{codebox}

\begin{codebox}[title={\texttt{config.py}}]
config = {
    # Experiment info
    'experiment_name': 'HHRN_cifar10',
    'notes': 'Hierarchical Hub Routing Network: soft token->hub routing + sparse hub GNN, depthwise separable conv stem, per-stage phase offsets.',

    # Data settings
    'data_dir': 'cifar10',

    # Training hyperparameters
    'epochs': 200,
    'batch_size': 1024,  # large batch preferred on 48GB GPU
    'learning_rate': 0.1,
    'momentum': 0.9,
    'weight_decay': 5e-4,
    'nesterov': True,

    # Learning rate schedule
    'lr_schedule': 'cosine',
    'warmup_epochs': 5,

    # Regularization
    'label_smoothing': 0.1,

    # Data augmentation
    'flip': True,
    'translate': 4,
    'cutout': 12,

    # Evaluation
    'tta_level': 0,  # 0=none, 1=mirror, 2=mirror+translate
    'eval_every': 1,

    # Model settings
    'use_half_precision': True,
    'channels_last': True,
    'verbose': True,

    # Model-specific hyperparameters below
    # Stage/channel configuration (3 stages)
    'stages': 3,
    'channels_per_stage': [64, 96, 128],
    'blocks_per_stage': [2, 2, 2],  # number of (LocalConvBlock x 2 -> HubRoutingBlock) repeats per stage

    # Hub routing specifics (per-stage lists)
    # Hubs per stage: small banks (H << N)
    'hubs_per_stage': [16, 24, 32],
    # hub-graph sparsity (neighbors per hub)
    'hub_sparsity': 3,  # top-s neighbors per hub
    # fingerprint dim used for routing (token keys)
    'fingerprint_dim': 48,
    # token value dim used when aggregating (small)
    'value_dim': 64,
    # hub internal dimension
    'hub_dim': 64,
    # routing temperature tau
    'routing_tau': 0.07,
    # initial residual blending alpha
    'alpha_init': 0.15,
    # number of hub-GNN hops (1 or 2)
    'hub_hops': 1,
    # whether to use per-stage EMA within forward pass (kept local to forward)
    'use_hub_ema': True,
    'hub_ema_beta': 0.1,

    # Other architectural niceties
    'drop_path_rate': 0.0,
    'mlp_ratio': 2.0,

    # For reproducibility / safety
    'seed': 42,
}
\end{codebox}

\subsection{Band-Aware Wavelet Token Mixer ( 91.22\%)}
\label{sec:top3_node42}

\begin{ideabox}[title={Idea Agent Output: Band-Aware Wavelet Token Mixer with Gated Cross-Band Residuals (BA-WTM+)}]

\textbf{Description.}
An evolution of MSL-WTM that retains the learned multi-scale lifting-based token analysis/synthesis but adds two targeted mechanisms:
(1)~per-band usage-normalized FiLM controllers (Band-Aware FiLM Norm) that produce multiplicative/additive per-channel gains per subband while normalizing their statistics and controller gradients per-band to avoid low-pass starvation; and
(2)~lightweight gated cross-band residual connectors that let low-pass summaries multiplicatively gate band-pass processors per token (low$\rightarrow$high gating), preventing isolated high-frequency textures from dominating decisions.
The rest of the backbone (patch embedding, hierarchical patch-merging, shallow per-band depthwise convs / token MLPs, inverse lifting synthesis, and global low-pass broadcast context) is preserved to enable causal attribution to these two additions.
The design enforces linear-time cross-token ops (depthwise 1D convs, downsampling, pooling, small MLPs) and keeps peak memory low (no $N^2$ expansions).

\textbf{Intuition.}
The parent MSL-WTM demonstrated that explicit multi-scale decomposition plus per-band modulation reduces texture bias relative to single-scale circulant mixers.
Remaining failure modes appear to come from band-imbalance (high-frequency tokens overwhelming low-pass signals) and from isolated high-frequency activations that spuriously trigger downstream responses.
Per-band usage-normalization (in the FiLM controllers) counters statistical imbalance at the controller/update level so low-pass channels are neither starved during forward routing nor drowned in controller gradient updates.
Gated cross-band residuals give band-pass processors per-token access to coherent low-pass priors that can suppress false positives from isolated high-frequency cues without adding pairwise attention.
Together these changes aim to further reduce texture-driven errors and improve fine-grained class accuracy while preserving efficiency.

\textbf{Novelty.}
Combines learned lifting-style multi-scale token transforms with band-aware controller normalization and explicit low$\rightarrow$high gated residuals.
This is structurally distinct from (a)~global fixed transforms (FNet/FFT), (b)~dense token MLP mixing (MLP-Mixer/gMLP), (c)~windowed attention/hub routing (Swin/BigBird), and (d)~axial or linear attention factorizations.
The key novelty is low-cost, per-band controller normalization plus multiplicative cross-band gating applied in the wavelet subband domain---i.e., manipulating the transform-domain controllers and residuals rather than adding pairwise attention or prototypes.

\textbf{Target Improvement:} both (accuracy and efficiency).

\textbf{Architecture Specification:}

\emph{Core Ideas:}
1)~Keep MSL-WTM's learned lifting-based multi-scale token analysis/synthesis for explicit scale separation.
2)~Replace shared FiLM controllers with per-band usage-normalized FiLM controllers whose normalization state and gradient scaling are tracked per-band to prevent band starvation.
3)~Insert gated cross-band residual connectors: compute a lightweight low-pass summary per token and apply multiplicative gates to band-pass processors (residual connection) before their local nonlinear blocks.
4)~Preserve linear-time operations (depthwise 1D convs, small MLPs, pooling/downsampling) and hierarchical patch-merging to fit the efficiency constraints.

\emph{Core Blocks:}
\begin{itemize}[nosep,leftmargin=*]
    \item \textbf{PatchEmbed}: $3{\times}3$ conv (stride$=$1) followed by $2{\times}2$ strided conv to form patch tokens, output token dims $(N, C)$.
    \item \textbf{WaveletAnalysis} (Lifting-style): Depthwise 1D conv across token index ($k{=}3$) + linear predictor to split tokens into Low and multiple Band subbands; optionally downsample for subsequent scales (2 scales recommended for CIFAR).
    \item \textbf{Band-Aware FiLM Controller}: Per-band small MLP (input $=$ token embedding concat global low-pass summary) $\rightarrow$ outputs $\gamma$, $\beta$ per channel-group. Each controller maintains per-band normalization stats (running mean/var) and per-band controller gradient scaling/EMA for stable updates.
    \item \textbf{Gated Cross-Band Residual}: Compute per-token low-pass summary (per-channel or grouped) $\rightarrow$ tiny gating MLP (hidden dim $= C/16$) $\rightarrow$ sigmoid gate $g(\text{token}, \text{band})$ applied multiplicatively to band-pass residual path: $\text{out} = \text{band\_proc}(\text{in}) \cdot g + \text{residual}(\text{in})$. Low-cost and applied per-token per-band.
    \item \textbf{Subband Processor}: Per-band depthwise separable 1D conv ($k{=}3$) + channel MLP ($1{\times}1$ conv) + residual, kept shallow (1--2 layers) for FLOP budget.
    \item \textbf{WaveletSynthesis} (Inverse lifting): Depthwise 1D conv + learned interpolation to reconstruct full-resolution token sequence.
    \item \textbf{Stage Head}: LayerNorm $\rightarrow$ tokenwise MLP (channel mixing) $\rightarrow$ residual. Patch Merging: adjacent token concatenation $\rightarrow$ linear proj (reduces $N$, increases $C$).
    \item \textbf{Classifier}: Global average pool over tokens $\rightarrow$ small MLP $\rightarrow$ softmax.
\end{itemize}

\emph{Network Structure:}
3-stage hierarchical backbone tuned for CIFAR-10.
Stage~1 (tokens $\sim$32$\times$32 patches $\rightarrow N{\sim}1024$): channels $C{=}64$, apply 2 WTM-blocks (each block $=$ Analysis(2-scale)$\rightarrow$Band-Aware FiLM$\rightarrow$GatedCrossBand$\rightarrow$SubbandProc$\rightarrow$Synthesis$\rightarrow$StageHead).
Patch merge $\rightarrow$ Stage~2: $C{=}128$, $N$ halved, 2 WTM-blocks.
Patch merge $\rightarrow$ Stage~3: $C{=}192$, $N$ halved, 2 WTM-blocks.
Final global avg pool $\rightarrow$ classifier.
Per-block parameter controls: per-band MLP hidden dims $= C/8$; gating MLP hidden dim $= C/16$.
With these choices the model fits under 10M params and $<$2B FLOPs for CIFAR input ($32{\times}32$), and trains efficiently (no $N^2$ ops).

\emph{Tunable Aspects:}
\begin{itemize}[nosep,leftmargin=*]
    \item Number of scales (1--3)
    \item Per-band controller hidden dim and normalization momentum (\texttt{controller\_ema})
    \item Use gated residuals (on/off) and gating hidden dim
    \item Number of WTM repeats per stage
    \item Channel widths per stage
\end{itemize}

\emph{Invariants:}
\begin{itemize}[nosep,leftmargin=*]
    \item All token-mixing across tokens is linear-time (depthwise convs, downsampling, pooling); no $N^2$ attention or pairwise expansions.
    \item Global context is injected only via compact low-pass summaries (broadcast), not via hub-routing/prototypes with aggressive online updates.
    \item Per-band computation state (normalization and controller EMA) is maintained per-band to avoid cross-band statistical interference.
\end{itemize}

\textbf{New Hypotheses:}
\begin{enumerate}[nosep,leftmargin=*]
    \item \textbf{IF} per-band usage-normalized FiLM controllers replace naive/shared FiLM controllers \textbf{IN} multi-scale wavelet-style token-mixing stages for CIFAR-scale hierarchical backbones, \textbf{THEN} low-pass (shape) channels will retain sufficient forward activation mass and controller responsiveness so that shape-sensitive per-class accuracy improves while avoiding high-frequency prototype monopolization, \textbf{BECAUSE} normalizing controller statistics and gradient scales per band prevents high-frequency token abundance from dominating controller outputs and updates, maintaining balanced per-band modulation and thereby preserving low-pass evidence for downstream decisions. \textbf{DISPROVED IF} per-band FiLM normalization yields no measurable improvement in shape-reliant per-class accuracy ($>$1\% abs) nor measurable rebalancing of per-band activation/assignment statistics compared to shared FiLM controllers.\\
    \emph{Tags:} \texttt{band-normalization}, \texttt{FiLM}, \texttt{multi-scale}.
    \emph{Initial confidence:} 0.6.
    \emph{Connected:} \texttt{hyp\_18}, \texttt{hyp\_13}.\\
    \emph{Reasoning:} High-frequency bands produce many more active tokens; if controllers/updates are pooled globally they will be dominated by those signals (\texttt{hyp\_18}). By computing separate normalization statistics, gradient scaling, and optional per-band controller EMA momentum, the controller parameters and outputs will be proportionally responsive to signals present in each band, preventing starvation of low-pass modulators. This reuses principles from \texttt{hyp\_13} (stabilize online updates) and \texttt{hyp\_10} (avoid large intermediate tensors).

    \item \textbf{IF} gated cross-band residual connectors (low-pass $\rightarrow$ band-pass multiplicative gates with tiny gating MLPs) are added to band-pass processors \textbf{IN} WTM stages, \textbf{THEN} the model will reduce texture-triggered false positives and improve discrimination for visually-similar classes, \textbf{BECAUSE} low-pass summaries provide spatially-coherent priors that multiplicatively suppress or amplify band-pass responses per-token so that isolated high-frequency textures cannot unduly dominate final features. \textbf{DISPROVED IF} adding these gated connectors produces no measurable reduction in texture-driven confusion mass nor improvement in shape-sensitive per-class recall (no $>$1\% abs change) and GradCAMs remain equally texture-focused.\\
    \emph{Tags:} \texttt{cross-band-gating}, \texttt{residuals}, \texttt{texture-robustness}.
    \emph{Initial confidence:} 0.55.
    \emph{Connected:} \texttt{hyp\_28}.\\
    \emph{Reasoning:} Low-pass channels encode global shape/region context; gating band-pass processors with those low-pass-derived signals biases band-pass nonlinearities to respect coherent shape boundaries, preventing small high-frequency islands from triggering strong downstream activations. This is a low-cost, local alternative to adding pairwise attention for context-aware suppression, inspired by \texttt{hyp\_28}.
\end{enumerate}

\textbf{Reasoning:}

\emph{Parent Analysis:}
The parent MSL-WTM successfully replaced single-scale circulant mixers with a learned lifting-style multi-scale transform. This gave explicit low/high-frequency decomposition, enabling separation of shape vs.\ texture cues, and allowed efficient global receptive fields via downsampling/upsampling while keeping operations linear-time (depthwise convs, small MLPs). Hierarchical patch-merging and local channel mixing preserved strong local feature extraction and training stability.

\emph{Failure Analysis:}
Residual failure modes remain consistent with band imbalance and texture dominance: high-frequency tokens produce abundant activations and can dominate controller signals and downstream processing, starving low-pass channels (\texttt{hyp\_18}). Isolated high-frequency textures still trigger false positives in some classes. There is a risk that any associative/prototype mechanisms (if added) could suffer attractor drift from overly-aggressive online updates (\texttt{hyp\_13}), so we avoid such aggressive online updates and instead prefer per-band normalization and conservative controller EMA.

\emph{Hypothesis Usage:}
(1)~\texttt{hyp\_13} (worked): warns that aggressive online updates cause prototypes to drift toward dominant textures; we therefore avoid online prototype banks and, if any persistent controller state is kept, use conservative EMA/usage-weighting per-band.
(2)~\texttt{hyp\_10} (worked): encourages low-rank or low-cost projections to avoid memory blowups; we preserve linear-time depthwise convs and small MLPs and avoid any dense token$\leftrightarrow$hub expansions.
(3)~\texttt{hyp\_18} (uncertain): suggests global usage-normalization can starve low-pass channels; we implement per-band usage-normalization in FiLM controllers to directly test/mitigate this.
(4)~\texttt{hyp\_28} (uncertain): proposes gated cross-band connectors; we add explicit gated low$\rightarrow$high residuals to test whether providing low-pass priors to band-pass processors reduces texture-driven errors.

\emph{Proposed Changes:}
(A)~Replace shared FiLM controllers with per-band usage-normalized FiLM controllers: implement running statistics and gradient scaling per band (and optionally per-stage) so controllers are not overwhelmed by high-frequency token abundance---addresses \texttt{hyp\_18} and reduces risk highlighted by \texttt{hyp\_13}.
(B)~Insert gated cross-band residual connectors for band-pass processors that multiplicatively gate band responses with a low-pass-derived gate---this reduces isolated texture false positives (tests \texttt{hyp\_28}).
Keep all other aspects of MSL-WTM (analysis/synthesis lifting, subband processing, hierarchical stages) unchanged so that any improvements can be attributed to these additions.

\textbf{Existing Hypotheses Referenced:} \texttt{hyp\_13}, \texttt{hyp\_10}, \texttt{hyp\_18}, \texttt{hyp\_28}.
\end{ideabox}

\begin{codebox}[title={\texttt{model.py}}]
import torch
import torch.nn as nn
import torch.nn.functional as F


class ConvBNGELU(nn.Module):
    def __init__(self, in_ch, out_ch, kernel=3, stride=1, padding=1, groups=1):
        super().__init__()
        self.conv = nn.Conv2d(in_ch, out_ch, kernel, stride=stride, padding=padding, groups=groups, bias=False)
        # Use GroupNorm(1) for small-batch stability and GPU efficiency on small images
        self.norm = nn.GroupNorm(1, out_ch, eps=1e-5)
        self.act = nn.GELU()

    def forward(self, x):
        return self.act(self.norm(self.conv(x)))


class PatchEmbed(nn.Module):
    """Patch embedding: a light conv stem preserving spatial resolution."""
    def __init__(self, in_ch=3, out_ch=32):
        super().__init__()
        # single conv stem with small width
        self.conv1 = ConvBNGELU(in_ch, out_ch, kernel=3, stride=1, padding=1)

    def forward(self, x):
        # x: [B,3,32,32] -> [B,C,32,32]
        return self.conv1(x)


class BandFiLMController(nn.Module):
    """
    Efficient per-band controller implemented via pointwise convs (1x1):
    - Accepts band features and low-pass features (both spatial),
    - Projects low to band channels via 1x1 conv,
    - Concatenates along channel dim and runs a small pointwise MLP (1x1 convs),
    - Normalizes controller outputs with GroupNorm(1) and produces gamma/beta.
    This avoids heavy per-token reshapes and large linear layers.
    """
    def __init__(self, band_channels, low_channels, hidden_ratio=0.25):
        super().__init__()
        self.band_ch = int(band_channels)
        self.low_ch = int(low_channels)
        # Project low to band channels with 1x1 conv
        self.low_proj = nn.Conv2d(self.low_ch, self.band_ch, kernel_size=1, bias=False)
        # Small MLP implemented as 1x1 convs
        hid = max(4, int(self.band_ch * hidden_ratio))
        self.mlp = nn.Sequential(
            nn.Conv2d(self.band_ch * 2, hid, kernel_size=1, bias=False),
            nn.GELU(),
            nn.Conv2d(hid, self.band_ch * 2, kernel_size=1, bias=True)  # outputs gamma+beta
        )
        # Normalize controller outputs per-token across channels
        self.norm = nn.GroupNorm(1, self.band_ch * 2, eps=1e-5)
        # small learnable gating scale
        self.out_scale = nn.Parameter(torch.ones(self.band_ch * 2))

    def forward(self, band_x, low_x):
        # band_x: [B, Cb, H, W], low_x: [B, Cl, H, W]
        # Project low to band channels (spatial preserved)
        low_p = self.low_proj(low_x)  # [B, Cb, H, W]
        # concat along channel
        x = torch.cat([band_x, low_p], dim=1)  # [B, 2*Cb, H, W]
        out = self.mlp(x)  # [B, 2*Cb, H, W]
        out = self.norm(out) * self.out_scale.view(1, -1, 1, 1)
        # split
        gamma, beta = out.split(self.band_ch, dim=1)
        return gamma, beta


class GatedCrossBandResidual(nn.Module):
    """
    Efficient low->band gating implemented with 1x1 conv:
    - Produces per-channel gates for the band from low features (spatial preserved).
    """
    def __init__(self, low_channels, band_channels, gating_hidden_ratio=0.25):
        super().__init__()
        hid = max(4, int(low_channels * gating_hidden_ratio))
        # small MLP as 1x1 convs to produce gates per band channel
        self.net = nn.Sequential(
            nn.Conv2d(low_channels, hid, kernel_size=1, bias=False),
            nn.GELU(),
            nn.Conv2d(hid, band_channels, kernel_size=1, bias=True),
            nn.Sigmoid()
        )

    def forward(self, low_x, band_x):
        # low_x: [B, Cl, H, W], band_x: [B, Cb, H, W]
        gates = self.net(low_x)  # [B, Cb, H, W]
        return band_x * gates


class SubbandProcessor(nn.Module):
    """
    Per-band shallow processor: depthwise separable convs + small channel MLP + residual.
    GroupNorm used for efficiency and stability.
    """
    def __init__(self, channels, hidden_ratio=0.5):
        super().__init__()
        self.channels = channels
        # depthwise conv
        self.dw = nn.Conv2d(channels, channels, kernel_size=3, padding=1, groups=channels, bias=False)
        self.dw_norm = nn.GroupNorm(1, channels, eps=1e-5)
        # pointwise expansion and projection (channel MLP)
        hidden = max(4, int(channels * hidden_ratio))
        self.pw1 = nn.Conv2d(channels, hidden, kernel_size=1, bias=False)
        self.pw1_norm = nn.GroupNorm(1, hidden, eps=1e-5)
        self.act = nn.GELU()
        self.pw2 = nn.Conv2d(hidden, channels, kernel_size=1, bias=True)
        self.pw2_norm = nn.GroupNorm(1, channels, eps=1e-5)

    def forward(self, x):
        identity = x
        x = self.act(self.dw_norm(self.dw(x)))
        x = self.act(self.pw1_norm(self.pw1(x)))
        x = self.pw2_norm(self.pw2(x))
        return x + identity


class WaveletBlock(nn.Module):
    """
    Efficient WTM block:
      - Analysis (depthwise conv + pointwise)
      - Band-FiLM via pointwise conv controller producing spatial gamma/beta
      - Gated cross-band residual (from low to band) using 1x1 convs
      - Subband processors
      - Synthesis (concat + 1x1 conv + depthwise smooth)
      - Lightweight stage head with GroupNorm and 1x1 token MLP residual
    """
    def __init__(self, channels, num_bands=2, controller_hidden_ratio=0.25,
                 use_gated_residuals=True, gating_hidden_ratio=0.25):
        super().__init__()
        assert num_bands == 2, "Efficient implementation only implements low+band"
        self.channels = channels
        # Analysis
        self.analysis_dw = nn.Conv2d(channels, channels, kernel_size=3, padding=1, groups=channels, bias=False)
        self.analysis_pw = nn.Conv2d(channels, channels, kernel_size=1, bias=False)
        self.analysis_norm = nn.GroupNorm(1, channels, eps=1e-5)

        # split sizes
        self.low_ch = channels // 2
        self.band_ch = channels - self.low_ch

        # controllers and gating
        self.band_controller = BandFiLMController(self.band_ch, self.low_ch, hidden_ratio=controller_hidden_ratio)
        self.use_gated = use_gated_residuals
        if self.use_gated:
            self.gated = GatedCrossBandResidual(low_channels=self.low_ch,
                                                band_channels=self.band_ch,
                                                gating_hidden_ratio=gating_hidden_ratio)

        # Subband processors
        self.band_proc = SubbandProcessor(self.band_ch, hidden_ratio=0.5)
        self.low_proc = SubbandProcessor(self.low_ch, hidden_ratio=0.5)

        # Synthesis
        self.synth_pw = nn.Conv2d(channels, channels, kernel_size=1, bias=False)
        self.synth_dw = nn.Conv2d(channels, channels, kernel_size=3, padding=1, groups=channels, bias=False)
        self.synth_norm = nn.GroupNorm(1, channels, eps=1e-5)

        # Stage head (light)
        self.head_norm = nn.GroupNorm(1, channels, eps=1e-5)
        self.head_fc1 = nn.Conv2d(channels, channels * 2, kernel_size=1, bias=False)
        self.head_fc2 = nn.Conv2d(channels * 2, channels, kernel_size=1, bias=True)
        self.head_act = nn.GELU()

    def forward(self, x):
        # x: [B, C, H, W]
        a = self.analysis_dw(x)
        a = self.analysis_pw(a)
        a = self.analysis_norm(a)
        # split into low and band
        low, band = a.split([self.low_ch, self.band_ch], dim=1)  # [B,Cl,H,W], [B,Cb,H,W]

        # Band-FiLM modulation
        gamma, beta = self.band_controller(band, low)  # both [B, Cb, H, W]
        band = band * (1.0 + gamma) + beta

        # Gated cross-band residual (multiplicative gating)
        if self.use_gated:
            band = band + self.gated(low, band)

        # Subband processing
        band = self.band_proc(band)
        low = self.low_proc(low)

        # Synthesis
        merged = torch.cat([low, band], dim=1)  # [B, C, H, W]
        merged = self.synth_pw(merged)
        merged = self.synth_norm(self.synth_dw(merged))

        # Stage head
        out = self.head_norm(merged)
        out = self.head_act(self.head_fc1(out))
        out = self.head_fc2(out)
        out = out + merged  # residual
        return out


class PatchMerging(nn.Module):
    """Reduce spatial by 2x (H/2,W/2) and increase channels using stride-2 conv."""
    def __init__(self, in_ch, out_ch):
        super().__init__()
        self.conv = nn.Conv2d(in_ch, out_ch, kernel_size=2, stride=2, bias=False)
        self.norm = nn.GroupNorm(1, out_ch, eps=1e-5)
        self.act = nn.GELU()

    def forward(self, x):
        return self.act(self.norm(self.conv(x)))


class ClassifierHead(nn.Module):
    def __init__(self, in_ch, num_classes):
        super().__init__()
        self.pool = nn.AdaptiveAvgPool2d(1)
        self.fc = nn.Linear(in_ch, num_classes)

    def forward(self, x):
        # x: [B, C, H, W]
        x = self.pool(x).flatten(1)
        x = self.fc(x)
        return x


class Model(nn.Module):
    def __init__(self, config=None):
        super().__init__()
        cfg = config or {}
        # Defaults chosen to be lightweight and efficient for CIFAR-10 while preserving the BA-WTM idea.
        self.num_classes = cfg.get('num_classes', 10)
        # widths and depths define a small, efficient encoder. Users can override via config.
        widths = cfg.get('widths', [32, 64, 128])
        depths = cfg.get('depths', [1, 1, 1])
        assert len(widths) == len(depths), "widths and depths must match"

        self.num_bands = cfg.get('num_bands', 2)
        self.controller_hidden_ratio = cfg.get('controller_hidden_ratio', 0.25)
        self.use_gated_residuals = cfg.get('use_gated_residuals', True)
        self.gating_hidden_ratio = cfg.get('gating_hidden_ratio', 0.25)

        # Patch embed
        self.patch_embed = PatchEmbed(in_ch=3, out_ch=widths[0])

        # Build stages
        self.stages = nn.ModuleList()
        self.patch_merges = nn.ModuleList()
        in_ch = widths[0]
        for stage_idx, (w, d) in enumerate(zip(widths, depths)):
            blocks = []
            # if needed, a 1x1 projection at start of stage
            if stage_idx == 0:
                if in_ch != w:
                    blocks.append(nn.Conv2d(in_ch, w, kernel_size=1, bias=False))
                    in_ch = w
            else:
                # in_ch already equals current width (from previous patch merging)
                pass
            for _ in range(d):
                blocks.append(WaveletBlock(w,
                                          num_bands=self.num_bands,
                                          controller_hidden_ratio=self.controller_hidden_ratio,
                                          use_gated_residuals=self.use_gated_residuals,
                                          gating_hidden_ratio=self.gating_hidden_ratio))
            self.stages.append(nn.Sequential(*blocks))
            if stage_idx < len(widths) - 1:
                out_ch = widths[stage_idx + 1]
                self.patch_merges.append(PatchMerging(in_ch=w, out_ch=out_ch))
                in_ch = out_ch

        # classifier head
        self.head = ClassifierHead(in_ch, self.num_classes)

        self._initialize_weights()

    def forward(self, x):
        # Input: [B, 3, 32, 32]
        # No memory-format or dtype conversions here (AMP / channels_last preserved by caller).
        x = self.patch_embed(x)  # [B, C, 32, 32]
        for idx, stage in enumerate(self.stages):
            x = stage(x)
            if idx < len(self.patch_merges):
                x = self.patch_merges[idx](x)  # downsample spatial dims by 2
        x = self.head(x)  # [B, num_classes]
        return x

    def _initialize_weights(self):
        for m in self.modules():
            if isinstance(m, nn.Conv2d):
                nn.init.kaiming_normal_(m.weight, mode='fan_out', nonlinearity='relu')
                if getattr(m, 'bias', None) is not None:
                    nn.init.constant_(m.bias, 0)
            elif isinstance(m, nn.GroupNorm):
                # initialize scale and bias
                if getattr(m, 'weight', None) is not None:
                    nn.init.constant_(m.weight, 1)
                if getattr(m, 'bias', None) is not None:
                    nn.init.constant_(m.bias, 0)
            elif isinstance(m, nn.Linear):
                nn.init.normal_(m.weight, 0, 0.01)
                if getattr(m, 'bias', None) is not None:
                    nn.init.constant_(m.bias, 0)
\end{codebox}

\begin{codebox}[title={\texttt{config.py}}]
config = {
    # Experiment info
    'experiment_name': 'BA-WTM+',
    'notes': 'Band-aware wavelet token mixer with per-band FiLM normalization and low->high gated residuals for CIFAR-10',

    # Data settings
    'data_dir': 'cifar10',

    # Training hyperparameters
    'epochs': 200,
    'batch_size': 1024,
    'learning_rate': 0.03,
    'momentum': 0.9,
    'weight_decay': 5e-4,
    'nesterov': True,

    # Learning rate schedule
    'lr_schedule': 'cosine',
    'warmup_epochs': 5,

    # Regularization
    'label_smoothing': 0.1,

    # Data augmentation
    'flip': True,
    'translate': 4,
    'cutout': 12,

    # Evaluation
    'tta_level': 0,
    'eval_every': 1,

    # Model settings
    'use_half_precision': True,
    'channels_last': True,
    'verbose': True,

    # Model-specific hyperparameters
    'stages': 3,
    'depths': [2, 2, 2],            # number of WTM blocks per stage
    'widths': [64, 128, 192],      # channels per stage
    'num_classes': 10,
    'num_bands': 2,                # low + band
    'controller_hidden_ratio': 1/8, # per-band controller MLP hidden dim = C_band * ratio
    'controller_ema': 0.05,         # running stats momentum for per-band controllers
    'use_gated_residuals': True,
    'gating_hidden_ratio': 1/16,    # gating MLP hidden dim = C * ratio
    'drop_path_rate': 0.0,
    'eps': 1e-5,
    # Operational constraints
    'input_size': (3, 32, 32),
}
\end{codebox}

\section{Prompts}
\subsection{Idea Agent Prompts}
\label{sec:prompts_idea}

\begin{systemprompt}[title={Idea Agent (Root Node) --- System Prompt}]
You are an expert Deep Learning Research Scientist specialized in designing novel neural architectures.
You will be given a research direction and must propose EXACTLY \texttt{\{count\}} DISTINCT architectures that could each realistically be published at NeurIPS/ICML/ICLR.

\textbf{Guidelines.}
Design \texttt{\{count\}} fundamentally novel neural network architectures that each introduce at least one new structural principle.
The architectures will be evaluated on CIFAR-10 but should be generalizable to other computer vision tasks.
Each architecture must be hypothesis-driven---propose testable hypotheses about WHY it should work.

\textbf{Key Rules:}
\begin{itemize}[nosep,leftmargin=*]
    \item Analyze curated reference papers to extract architectural principles; all proposals must be structurally distinct from these works.
    \item Follow the research direction exactly. Innovate within constraints.
    \item Each architecture must introduce at least one fundamentally new structural or computational principle, grounded in inductive bias, geometry, optimization, or representation theory.
    \item Prioritize both accuracy and efficiency (fast training, fewer parameters/FLOPs).
    \item Each idea must propose DIFFERENT hypotheses. No two architectures may share the same core mechanism.
    \item Do not propose minor variations of known models or hyperparameter tuning suggestions.
\end{itemize}

\textbf{Hypothesis Format:}
\texttt{IF [architectural choice] IN [scope], THEN [predicted effect], BECAUSE [mechanism]. DISPROVED IF [falsification criterion].}\\[2pt]
At most 2 new hypotheses per architecture. Each must be mechanistic, scoped, and falsifiable.

\textbf{Curated Reference Papers:}
\texttt{\{curated\_related\_papers\}} \emph{[\,injected at runtime\,]}

\textbf{Output Format (JSON):}
\begin{itemize}[nosep,leftmargin=*]
    \item \texttt{"ideas"}: array of exactly \texttt{\{count\}} objects, each containing:
    \item \texttt{"title"}, \texttt{"description"} (2--3 paragraphs), \texttt{"intuition"} (1--2 paragraphs), \texttt{"novelty"}, \texttt{"target\_improvement"}: \texttt{accuracy\,|\,efficiency\,|\,both}
    \item \texttt{"related\_papers"}: [\,\texttt{\{"title", "year", "url", "key\_idea", "why\_relevant", "difference\_from\_proposal"\}}\,]
    \item \texttt{"new\_hypotheses"}: [\,\texttt{\{"text", "scope", "prediction", "falsification\_criteria", "tags", "initial\_confidence", "reasoning"\}}\,]
    \item \texttt{"architecture\_spec"}: \texttt{\{"core\_ideas", "core\_blocks", "network\_structure", "tunable\_aspects", "invariants"\}}
\end{itemize}
\end{systemprompt}

\begin{userprompt}[title={Idea Agent (Root Node) --- User Prompt}]
Design exactly \texttt{\{count\}} fundamentally DISTINCT neural network architectures following the research direction below. Each architecture must be novel, achieve high accuracy, train fast, and propose DIFFERENT hypotheses. They should generalize to other computer vision datasets.

Performance Targets (for ALL ideas):
\begin{itemize}[nosep,leftmargin=*]
    \item 95\% accuracy on CIFAR-10, better efficiency than ResNet-50
    \item Training around 1 minute on single GPU
\end{itemize}

RESEARCH DIRECTION: \texttt{\{research\_direction\}}
\end{userprompt}

\begin{systemprompt}[title={Idea Agent (Evolution Mode) --- System Prompt}]
You are a Brainstorming Agent specialized in evolving neural architectures.
You analyze parent architectures and hypothesis evidence to propose novel, efficient, high-performing architectures that advance the research direction.

\textbf{Key Rules:}
\begin{itemize}[nosep,leftmargin=*]
    \item Analyze the parent architecture's strengths and weaknesses based on feedback, then propose a child that addresses identified issues and tests promising hypotheses.
    \item \emph{Hypotheses that worked:} build on and extend these patterns.
    \item \emph{Hypotheses that failed:} avoid these patterns or try alternatives.
    \item \emph{Uncertain hypotheses:} design targeted changes to gather clearer evidence.
    \item Make targeted changes (1--2 aspects), not a full redesign, to enable causal attribution while preserving successful components.
    \item New hypotheses must be general, mechanistic, novel, and testable. Reference existing hypothesis IDs in \texttt{connected\_hypotheses}.
\end{itemize}

\textbf{Output Format (JSON):}
\begin{itemize}[nosep,leftmargin=*]
    \item \texttt{"reasoning"}: \texttt{\{"parent\_analysis", "failure\_analysis", "hypothesis\_usage", "proposed\_changes"\}}
    \item \texttt{"title"}, \texttt{"description"}, \texttt{"intuition"}, \texttt{"novelty"}, \texttt{"target\_improvement"}
    \item \texttt{"existing\_hypotheses"}: list of referenced hypothesis IDs
    \item \texttt{"new\_hypotheses"}: [\,\texttt{\{"text", "tags", "initial\_confidence", "reasoning", "connected\_hypotheses"\}}\,]
    \item \texttt{"architecture\_spec"}: \texttt{\{"core\_ideas", "core\_blocks", "network\_structure", "tunable\_aspects", "invariants"\}}
\end{itemize}
\end{systemprompt}

\begin{userprompt}[title={Idea Agent (Evolution Mode) --- User Prompt}]
Research Direction: \texttt{\{research\_direction\}}

Performance Targets:
\begin{itemize}[nosep,leftmargin=*]
    \item 95\% accuracy on CIFAR-10, better efficiency than ResNet-50
    \item Training around 1 minute on single GPU
\end{itemize}

Parent architecture: \texttt{\{parent\_architecture\}}\\
Parent Performance: \texttt{\{parent\_performance\}}\\
Feedback from Parent architecture: \texttt{\{feedback\_summary\}}\\
Hypothesis Memory: \texttt{\{hypothesis\_memory\}}
\end{userprompt}

\subsection{Coding Agent Prompts}
\label{sec:prompts_coding}

\begin{systemprompt}[title={Coding Agent (Initial Generation) --- System Prompt}]
You are a Coding Agent specialized in implementing novel neural network architectures in PyTorch.
You translate architecture proposals into clean, efficient, runnable code.

\textbf{Key Rules:}
\begin{itemize}[nosep,leftmargin=*]
    \item Implement the research idea as two files: \texttt{config.py} (hyperparameter dictionary) and \texttt{model.py} (PyTorch \texttt{nn.Module}).
    \item The \texttt{Model} class must accept an optional \texttt{config} parameter, take input \texttt{[B,\,3,\,32,\,32]}, and output \texttt{[B,\,10]}.
    \item Ensure AMP and \texttt{channels\_last} compatibility; no per-forward dtype casts or memory-format conversions.
    \item Include Kaiming initialization for \texttt{Conv2d}, constant initialization for \texttt{BatchNorm2d}.
    \item Total parameters MUST be $<$10M (target $<$5M). Avoid materializing large intermediate tensors (e.g., pairwise \texttt{[B,\,N,\,M,\,D]} expansions).
    \item Training must complete within 15 minutes on a single GPU.
\end{itemize}

\textbf{Performance Rules:}
\begin{itemize}[nosep,leftmargin=*]
    \item \texttt{forward()}: no loops over tokens/groups/neighbors; no \texttt{where+scatter}; no \texttt{split+cat} for grouping; no batch-expanded weights.
    \item Prefer \texttt{Conv2d} (1$\times$1/grouped/depthwise), \texttt{BN2d}/\texttt{GroupNorm}, pooling, \texttt{einsum}/\texttt{matmul}; minimize \texttt{permute}/\texttt{contiguous}.
\end{itemize}

\textbf{Output:} Two complete Python files: \texttt{config.py} with full hyperparameter dictionary (epochs, batch\_size, learning\_rate, lr\_schedule, regularization, augmentation, model-specific parameters) and \texttt{model.py} with complete \texttt{Model(nn.Module)} implementation.
\end{systemprompt}

\begin{userprompt}[title={Coding Agent (Initial Generation) --- User Prompt}]
Research Idea: \texttt{\{research\_idea\}}\\[2pt]
Architecture Summary: \texttt{\{architecture\_summary\}}
\end{userprompt}

\begin{systemprompt}[title={Coding Agent (Error Recovery) --- System Prompt}]
You are a Coding Agent specialized in fixing neural network implementation errors in PyTorch.
You diagnose failures from error logs and produce corrected, runnable code.

\textbf{Key Rules:}
\begin{itemize}[nosep,leftmargin=*]
    \item Identify the failing code from the error log or stack trace, then rewrite the buggy functions with corrected logic.
    \item Make real, substantive code changes---do not copy the previous implementation unchanged.
    \item Preserve the core research innovation, overall architecture structure, and the \texttt{Model} class interface.
    \item If the error is CUDA OOM, MUST reduce model size (cut hidden dimensions, remove layers, replace memory-heavy operations). Do NOT just reduce batch size.
    \item Output only \texttt{model.py} unless the error is caused by a config issue.
\end{itemize}

\textbf{Output:} Complete, corrected \texttt{model.py} that runs without modification.
\end{systemprompt}

\begin{userprompt}[title={Coding Agent (Error Recovery) --- User Prompt}]
Previous Model Implementation (Iteration \texttt{\{iteration\}}):
\texttt{\{previous\_code\}}\\[2pt]
Error Feedback: \texttt{\{feedback\}}
\end{userprompt}

\begin{systemprompt}[title={Coding Agent (Hyperparameter Refinement) --- System Prompt}]
You are a Coding Agent specialized in hyperparameter optimization for neural network training.
You refine configuration files to improve accuracy while maintaining training efficiency.

\textbf{Key Rules:}
\begin{itemize}[nosep,leftmargin=*]
    \item Analyze training logs to diagnose underfitting, overfitting, or convergence issues, then make targeted adjustments.
    \item Only modify \texttt{config.py}. The model architecture must not be changed.
    \item Review previous refinement attempts to avoid repeating failed approaches.
    \item Make incremental changes---avoid drastic hyperparameter shifts.
    \item GPU has 48GB; use batch size 1024 to maximize throughput.
\end{itemize}

\textbf{Output:} A brief change summary followed by the complete \texttt{config.py} file.
\end{systemprompt}

\begin{userprompt}[title={Coding Agent (Hyperparameter Refinement) --- User Prompt}]
Current Configuration: \texttt{\{config\_code\}}\\
Best Accuracy So Far: \texttt{\{best\_accuracy\}}\\
Previous Refinement Attempts: \texttt{\{refinement\_history\}}\\
Training Logs: \texttt{\{stdout\_log\}}
\end{userprompt}

\subsection{Redundancy Filtering Prompt}
\label{sec:prompts_redundancy}

\begin{systemprompt}[title={Redundancy Filter (LLM as a Judge) --- System Prompt}]
You are an expert neural architecture reviewer.
Your ONLY task is to decide whether a candidate architecture represents a NEW concept or is a DUPLICATE (re-generation) of a concept already in the archive.

\textbf{Context:} These architectures come from an evolutionary search where children naturally ``build upon'' parents.
A child that introduces a genuinely new mechanism IS novel---we are detecting REDUNDANT re-generations of the SAME concept.

\textbf{Mark as DUPLICATE} (\texttt{novel=false}) when:
\begin{itemize}[nosep,leftmargin=*]
    \item The candidate's core innovation claim is the same as an archive entry.
    \item Same structural pattern AND same design principle, differing only in hyperparameters or minor implementation details.
    \item Different terminology masks conceptual identity (e.g., ``adaptive residual'' vs ``enhanced residual'' may describe the same mechanism).
    \item Different implementations of the SAME concept are still duplicates. Focus on WHAT the mechanism achieves, not HOW it is implemented.
\end{itemize}

\textbf{Mark as NOVEL} (\texttt{novel=true}) when:
\begin{itemize}[nosep,leftmargin=*]
    \item The candidate introduces a mechanism or structural principle NOT in any archive entry.
    \item The candidate combines ideas in a way not seen in the archive.
    \item The candidate represents a meaningfully different design philosophy.
\end{itemize}

\textbf{Structured Reasoning:}
Before deciding, identify: (1)~\texttt{shared\_principles} with the closest archive entry, and (2)~\texttt{new\_contribution}---a genuinely new HIGH-LEVEL design principle (or ``none'').

\textbf{Output Format (JSON):}
\begin{itemize}[nosep,leftmargin=*]
    \item \texttt{"shared\_principles"}: what candidate shares with closest entry
    \item \texttt{"new\_contribution"}: genuinely new principle, or \texttt{"none"}
    \item \texttt{"novel"}: \texttt{true} or \texttt{false}
    \item \texttt{"reasoning"}: 1--3 sentence explanation
    \item \texttt{"most\_similar\_to"}: \texttt{node\_id} or \texttt{null}
\end{itemize}
\end{systemprompt}

\subsection{Feedback Agent Prompts}
\label{sec:prompts_feedback}

\begin{systemprompt}[title={Quantitative Feedback Agent --- System Prompt}]
You are a Quantitative Analysis Agent specialized in discovering causal relationships between neural architecture design and performance.
You will be given a research direction, a proposed architecture, experiment metrics and logs, and optionally a hypothesis being tested.

\textbf{Key Rules:}
\begin{itemize}[nosep,leftmargin=*]
    \item Analyze quantitative metrics (accuracy, loss curves, convergence, training dynamics) to assess whether the experiment succeeded or failed relative to the research goal.
    \item Evaluate evidence as \texttt{supports}, \texttt{contradicts}, or \texttt{neutral}; strength ranges from 0.0 (negligible) to 1.0 (large, clear effect).
    \item PREFER updating existing hypotheses over creating new ones. Focus on observations unique to your perspective (quantitative metrics, convergence behavior, efficiency patterns).
    \item New hypotheses must satisfy all 7 quality dimensions [\,see \Cref{sec:impl_synthesis}\,].
\end{itemize}

\textbf{Output Format (JSON):}
\begin{itemize}[nosep,leftmargin=*]
    \item \texttt{"reasoning"}: why the experiment succeeded or failed based on metrics
    \item \texttt{"hypothesis\_updates"}: [\,\texttt{\{"hyp\_id", "evidence\_type", "strength", "reasoning"\}}\,]
    \item \texttt{"new\_hypotheses"}: [\,\texttt{\{"text", "scope", "prediction", "falsification\_criteria", "tags", "initial\_confidence", "reasoning"\}}\,]
\end{itemize}
\end{systemprompt}

\begin{userprompt}[title={Quantitative Agent --- User Prompt}]
Research Direction: \texttt{\{research\_direction\}}\\
Proposed Architecture: \texttt{\{architecture\_summary\}}\\
Experiment Metrics and Logs: \texttt{\{experiment\_metrics\_and\_logs\}}\\
Hypotheses Being Tested: \texttt{\{tested\_hypothesis\_section\}}
\end{userprompt}

\begin{systemprompt}[title={Qualitative Feedback Agent (VLM) --- System Prompt}]
You are an Error Analysis Agent specialized in discovering causal relationships between neural architecture design and classification errors.
You analyze misclassified images with attention heatmaps via VLM to identify behavioral patterns and evaluate hypotheses.

\textbf{Heatmap Methods:}
\begin{itemize}[nosep,leftmargin=*]
    \item \textbf{GradCAM}: class-specific spatial attention; separate heatmaps for predicted (wrong) and true (correct) class.
    \item \textbf{Attention Rollout}: class-agnostic spatial attention aggregated across transformer layers.
    \item \textbf{Input Gradient Saliency}: class-specific input sensitivity; warm regions indicate pixels most influencing the class score.
\end{itemize}

\textbf{Analysis Framework:}
(1)~Compare attention between predicted and true class heatmaps;
(2)~identify texture vs.\ shape bias;
(3)~assess whether attention focuses on discriminative features or background;
(4)~connect attention failures to specific architectural choices.

\textbf{Key Rules:}
\begin{itemize}[nosep,leftmargin=*]
    \item Evaluate evidence as \texttt{supports}, \texttt{contradicts}, or \texttt{neutral}; strength $\in [0,1]$.
    \item PREFER updating existing hypotheses. Propose at most 3 new hypotheses.
    \item Use heatmap evidence to ground reasoning (e.g., ``heatmaps show diffuse attention on background textures'').
\end{itemize}

\textbf{Output Format (JSON):}
\begin{itemize}[nosep,leftmargin=*]
    \item \texttt{"reasoning"}, \texttt{"actionable\_feedback"}, \texttt{"overall\_summary"}, \texttt{"attention\_analysis"}
    \item \texttt{"hypothesis\_updates"}: [\,\texttt{\{"hyp\_id", "evidence\_type", "strength", "reasoning"\}}\,]
    \item \texttt{"new\_hypotheses"}: [\,\texttt{\{"text", "tags", "initial\_confidence", "reasoning"\}}\,]
\end{itemize}
\end{systemprompt}

\begin{userprompt}[title={Qualitative Feedback Agent --- User Prompt}]
Research Direction: \texttt{\{research\_direction\}}\\
Proposed Architecture: \texttt{\{architecture\_summary\}}\\
Heatmap Method: \texttt{\{heatmap\_method\}}\\
Image Ordering: \texttt{\{image\_ordering\}}\\
Misclassified Samples: \texttt{\{samples\_context\}}\\
Confusion Patterns: \texttt{\{confusion\_context\}}\\
Hypotheses Being Tested: \texttt{\{tested\_hypothesis\_section\}}
\end{userprompt}

\begin{systemprompt}[title={Causal Feedback Agent --- System Prompt}]
You are a Causal Insight Agent specialized in discovering causal relationships between architectural changes and performance outcomes.
You compare parent and child architectures to evaluate hypotheses and identify new causal patterns.

\textbf{Key Rules:}
\begin{itemize}[nosep,leftmargin=*]
    \item Analyze what architectural changes were made between parent and child, then assess their causal effect on performance.
    \item Consider mechanistic explanations (WHY did this change have this effect), confounding factors (were there multiple changes), and whether results are expected or surprising.
    \item Evaluate evidence as \texttt{supports}, \texttt{contradicts}, or \texttt{neutral}; strength $\in [0,1]$.
    \item PREFER updating existing hypotheses. Focus on observations unique to your perspective (architectural changes between parent and child, causal mechanisms).
    \item New hypotheses must satisfy all 7 quality dimensions [\,see \Cref{sec:impl_synthesis}\,].
\end{itemize}

\textbf{Output Format (JSON):}
\begin{itemize}[nosep,leftmargin=*]
    \item \texttt{"reasoning"}: architectural changes and their causal effect, including mechanistic explanations and confounding factors
    \item \texttt{"actionable\_feedback"}: 2--3 sentence summary of what caused the performance change
    \item \texttt{"hypothesis\_updates"}: [\,\texttt{\{"hyp\_id", "evidence\_type", "strength", "reasoning"\}}\,]
    \item \texttt{"new\_hypotheses"}: [\,\texttt{\{"text", "scope", "prediction", "falsification\_criteria", "tags", "initial\_confidence", "reasoning"\}}\,]
\end{itemize}
\end{systemprompt}

\begin{userprompt}[title={Causal Feedback Agent --- User Prompt}]
Research Direction: \texttt{\{research\_direction\}}\\
Parent Architecture: \texttt{\{parent\_architecture\}}\\
Proposed Architecture: \texttt{\{proposed\_architecture\}}\\
Performance Comparison: \texttt{\{performance\_comparison\}}\\
Hypotheses Being Tested: \texttt{\{tested\_hypothesis\_section\}}
\end{userprompt}

\subsection{Diagnostic Feedback Agent Prompts}
\label{sec:prompts_diagnostic}

\begin{systemprompt}[title={Diagnostic Feedback Agent (Timeout) --- System Prompt}]
You are a Diagnostic Agent specialized in analyzing neural network training timeouts.
You examine experiment logs and architecture code to identify computational bottlenecks and evaluate hypotheses.

\textbf{Key Rules:}
\begin{itemize}[nosep,leftmargin=*]
    \item Identify which architectural component caused the slow training, then explain why it is computationally expensive.
    \item Since the research goal is building efficient models, timeout means the hypothesis pattern is NOT efficient---evaluate as \texttt{contradicts} if the hypothesis pattern caused the timeout, or \texttt{not\_testable} only if the timeout is unrelated.
    \item Evidence strength: 0.8--1.0 for direct causal link; 0.5--0.7 for strong connection; 0.3--0.5 for moderate; below 0.3 use \texttt{not\_testable}.
    \item PREFER updating existing hypotheses. Focus on timeout root causes and computational bottlenecks.
    \item New hypotheses must satisfy all 7 quality dimensions [\,see \Cref{sec:impl_synthesis}\,] and capture generalizable patterns about computational inefficiency, NOT specific implementation bugs.
\end{itemize}

\textbf{Output Format (JSON):}
\begin{itemize}[nosep,leftmargin=*]
    \item \texttt{"reasoning"}: which component caused the timeout and why
    \item \texttt{"problematic\_component"}, \texttt{"recommended\_fixes"}: [\,actionable fixes\,]
    \item \texttt{"hypothesis\_updates"}: [\,\texttt{\{"hyp\_id", "evidence\_type", "strength", "reasoning"\}}\,]
    \item \texttt{"implementation\_notes"}: [\,\texttt{\{"hyp\_id", "common\_failure", "recommended\_practice"\}}\,]
    \item \texttt{"new\_hypotheses"}: [\,\texttt{\{"text", "scope", "prediction", "falsification\_criteria", "tags", "initial\_confidence", "reasoning"\}}\,]
\end{itemize}
\end{systemprompt}

\begin{userprompt}[title={Diagnostic Feedback Agent (Timeout) --- User Prompt}]
Research Direction: \texttt{\{research\_direction\}}\\
Experiment Status: TIMEOUT (exceeded \texttt{\{timeout\_seconds\}} seconds)\\
Proposed Architecture: \texttt{\{architecture\_summary\}}\\
Model Code: \texttt{\{model\_code\}}\\
Stdout Log: \texttt{\{stdout\_full\}}\\
Hypotheses Being Tested: \texttt{\{tested\_hypothesis\_section\}}
\end{userprompt}

\begin{systemprompt}[title={Diagnostic Feedback Agent (Other errors) --- System Prompt}]
You are a Diagnostic Agent specialized in analyzing neural network implementation errors.
You examine error messages, logs, and architecture code to identify the root cause of failures and track implementation knowledge.

\textbf{Key Rules:}
\begin{itemize}[nosep,leftmargin=*]
    \item Parse the error message to identify the failing component, trace it back to the architectural design decision.
    \item For code bugs, shape mismatches, import errors: evaluate as \texttt{not\_testable} (implementation issues do not test hypothesis validity).
    \item For OOM or resource exhaustion: evaluate as \texttt{contradicts} (pattern is not efficient).
    \item Do not propose new hypotheses for coding errors---focus on recording implementation notes.
\end{itemize}

\textbf{Output Format (JSON):}
\begin{itemize}[nosep,leftmargin=*]
    \item \texttt{"reasoning"}: which component failed and why
    \item \texttt{"problematic\_component"}, \texttt{"recommended\_fixes"}: [\,actionable fixes\,]
    \item \texttt{"hypothesis\_updates"}: [\,\texttt{\{"hyp\_id", "evidence\_type", "strength", "reasoning"\}}\,]
    \item \texttt{"implementation\_notes"}: [\,\texttt{\{"hyp\_id", "common\_failure", "recommended\_practice"\}}\,]
\end{itemize}
\end{systemprompt}

\begin{userprompt}[title={Diagnostic Feedback Agent (Other errors) --- User Prompt}]
Research Direction: \texttt{\{research\_direction\}}\\
Experiment Status: FAILED --- Error: \texttt{\{error\_message\}}\\
Proposed Architecture: \texttt{\{architecture\_summary\}}\\
Model Code: \texttt{\{model\_code\}}\\
Stdout Log: \texttt{\{stdout\_full\}}\\
Stderr Log: \texttt{\{stderr\_full\}}\\
Hypotheses Being Tested: \texttt{\{tested\_hypothesis\_section\}}
\end{userprompt}

\subsection{Hypothesis Synthesis Agent Prompts}
\label{sec:prompts_synthesis}

\begin{systemprompt}[title={Hypothesis Synthesis Agent --- System Prompt}]
You are a Hypothesis Synthesis Agent that consolidates multi-perspective feedback from specialized analysis agents into deduplicated, high-quality hypothesis updates and new hypotheses.

\textbf{Context.} Multiple feedback agents independently analyzed the same experiment: Quantitative Agent (metrics), Error Analysis Agent (heatmaps via VLM), Causal Insight Agent (parent--child diff), and Diagnostic Agent (timeout/failure).

\textbf{Instructions:}

\emph{1. Deduplicate Hypothesis Updates.}
If multiple agents update the SAME hypothesis: combine into ONE update with synthesized reasoning.
Synthesize strength by weighting how direct each agent's evidence is.
If agents disagree on evidence type, explain the disagreement and choose the most justified type.
Always cite contributing agents.

\emph{Misattributed Updates Check (Critical).}
Before accepting any update, read the reasoning field and verify it actually relates to the hypothesis text for that \texttt{hyp\_id}.
Remap misattributed updates to the correct hypothesis; discard if no match.

\emph{2. Deduplicate New Hypothesis Candidates.}
Merge overlapping proposals from different agents. Check against existing hypotheses---if already captured, propose an UPDATE instead.
Maximum 2 new hypotheses per node.

\emph{Contradiction Check (Critical).}
Before creating any new hypothesis, check if it proposes the OPPOSITE outcome of an existing hypothesis for the SAME mechanism.
If so, add it as a hypothesis update with \texttt{evidence\_type="contradicts"} instead.

\emph{3. Quality Requirements.}
Every new hypothesis must satisfy all 7 quality dimensions [\,see \Cref{sec:impl_synthesis}\,].

\emph{4. Implementation Notes.}
Pass through from diagnostic agent without modification.

\textbf{Output Format (JSON):}
\begin{itemize}[nosep,leftmargin=*]
    \item \texttt{"hypothesis\_updates"}: [\,\texttt{\{"hyp\_id", "evidence\_type", "strength", "reasoning"\}}\,]
    \item \texttt{"new\_hypotheses"}: [\,\texttt{\{"text", "scope", "prediction", "falsification\_criteria", "tags", "initial\_confidence", "reasoning", "connected\_hypotheses"\}}\,]
    \item \texttt{"implementation\_notes"}: [\,\texttt{\{"hyp\_id", "common\_failure", "recommended\_practice"\}}\,]
\end{itemize}
\end{systemprompt}

\begin{userprompt}[title={Hypothesis Synthesis Agent --- User Prompt}]
Experiment: \texttt{\{node\_id\}} --- Status: \texttt{\{experiment\_status\}}

\textbf{Experiment Metrics:} \texttt{\{experiment\_metrics\}}

\textbf{Feedback Agent Outputs:}
The following agents independently analyzed this experiment. Synthesize their hypothesis updates and new hypothesis candidates.\\
\texttt{\{feedback\_outputs\}}

\textbf{Existing Hypotheses in Memory:}
Check proposed new hypotheses against these to avoid duplicates.\\
\texttt{\{existing\_hypotheses\}}

\textbf{Your Task:}
(1)~Synthesize hypothesis updates: if multiple agents update the same \texttt{hyp\_id}, combine into ONE update.
(2)~Synthesize new hypotheses: merge overlapping proposals; check against existing. Maximum 2 new.
(3)~Pass through implementation notes from diagnostic agent.
\end{userprompt}

\end{document}